\documentclass{article} % For LaTeX2e

\usepackage[nonatbib, preprint]{neurips_2020}

\usepackage[utf8]{inputenc} % allow utf-8 input
\usepackage[T1]{fontenc}    % use 8-bit T1 fonts
\usepackage{hyperref}       % hyperlinks
\usepackage{url}            % simple URL typesetting
\usepackage{booktabs}       % professional-quality tables
\usepackage{amsfonts}       % blackboard math symbols
\usepackage{nicefrac}       % compact symbols for 1/2, etc.
\usepackage{microtype}      % microtypography

\usepackage{graphicx}
\usepackage{subfigure}
\usepackage{multirow}
\usepackage{wrapfig}

\newcommand{\culeurl}{\url{https://github.com/NVlabs/cule}}

\begin{document}
\title{Accelerating Reinforcement Learning through GPU Atari Emulation}

% Authors must not appear in the submitted version. They should be hidden
% as long as the \iclrfinalcopy macro remains commented out below.
% Non-anonymous submissions will be rejected without review.

\author{Steven Dalton\thanks{Equal contribution}, Iuri Frosio\footnotemark[1], \& Michael Garland \\
NVIDIA, USA\\
\texttt{\{sdalton,ifrosio,mgarland\}@nvidia.com}
}

\maketitle

\begin{abstract}
We introduce CuLE (CUDA Learning Environment), a CUDA port of the Atari
Learning Environment (ALE) which is used for the development of deep
reinforcement algorithms.  CuLE overcomes many limitations of existing
CPU-based emulators and scales naturally to multiple GPUs.  It leverages GPU
parallelization to run thousands of games simultaneously and it renders frames
directly on the GPU, to avoid the bottleneck arising from the limited CPU-GPU
communication bandwidth.  CuLE generates up to 155M frames per hour on a single
GPU, a finding previously achieved only through a cluster of CPUs.  Beyond
highlighting the differences between CPU and GPU emulators in the context of
reinforcement learning, we show how to leverage the high throughput of CuLE by
effective batching of the training data, and show accelerated convergence for
A2C+V-trace. CuLE is available at \culeurl.

\end{abstract}

% \begin{IEEEkeywords}
% GPU, Deep reinforcement learning.
% \end{IEEEkeywords}

% \fi
% \IEEEpeerreviewmaketitle

\section{Introduction}
~\label{sec:introduction}

Initially triggered by the success of DQN~\cite{Mnih:2015}, research in Deep
Reinforcement Learning (DRL) has grown in popularity in the last
years~\cite{Lil15, Mnih:2016, Mnih:2015}, leading to intelligent agents that
solve non-trivial tasks in complex environments. But DRL also soon proved to be
a challenging computational problem, especially if one wants to achieve peak
performance on modern architectures.

% Fig.~\ref{fig:design} is a schematic representation of a modern DRL algorithm
% implementation:

Traditional DRL training focuses on CPU environments that execute a set of
actions $\{a_{t-1}\}$ at time $t-1$, and produce observable states $\{s_t\}$
and rewards $\{r_t\}$.  These data are migrated to a Deep Neural Network (DNN)
on the GPU to eventually select the next action set, $\{a_t\}$, which is copied
back to the CPU.  This sequence of operations defines the \emph{inference
path}, whose main aim is to generate training data.  A training buffer on the
GPU stores the states generated on the \emph{inference path}; this is
periodically used to update the DNN's weights $\theta$, according to the
training rule of the DRL algorithm (\emph{training path}).  A computationally
efficient DRL system should balance the data generation and training processes,
while minimizing the communication overhead along the \emph{inference path} and
consuming, along the \emph{training path}, as many data per second as
possible~\cite{Babaeizadeh:2016,Babaeizadeh:2017}.  The solution to this
problem is however non-trivial and many DRL implementations do not leverage the
full computational potential of modern systems~\cite{Stooke:2018}.

\begin{table}[ht!]
\caption{Average training times, raw frames to reach convergence, FPS, and
  computational resources of existing accelerated DRL schemes, compared to
  CuLE. Data from~\cite{Horgan:2018}; FPS are taken from the corresponding
  papers, if available, and measured on the entire Atari suite for CuLE.}
\label{tab:compute}
\centering
%\begin{tiny}
%\begin{sc}
\scalebox{0.65}{
\begin{tabular}{lccccc}
\toprule
Algorithm & Time & Frames & FPS & Resources \\
\midrule
  Ape-X DQN~\cite{Horgan:2018} & 5 days & 22,800M & 50K & 376 cores, 1 GPU &\\
  Rainbow~\cite{Hessel:2017} & 10 days & 200M & --- & 1 GPU &\\
  Distributional (C51)~\cite{Bellemare:2017}  & 10 days & 200M & --- & 1 GPU &\\
  A3C~\cite{Mnih:2016} & 4 days & --- & 2K & 16 cores &\\
  GA3C~\cite{Babaeizadeh:2016,Babaeizadeh:2017} & 1 day & --- & 8K & 16 cores, 1 GPU &\\
  Prioritized Dueling~\cite{Wang:2016} & 9.5 days & 200M & --- & 1 GPU &\\
  DQN~\cite{Mnih:2015} & 9.5 days & 200M & --- & 1 GPU &\\
\midrule
  Gorila DQN~\cite{Nair:2015} & ~4 days & --- & --- & $>100$ cores &\\
  Unreal~\cite{Jaderberg:2016} & --- & 250M & --- & 16 cores &\\
\midrule
  Stooke (A2C / DQN)~\cite{Stooke:2018} & hours & 200M & 35K & 40 CPUs, 8 GPUs (DGX-1) &\\
  IMPALA (A2C~+~V-Trace)~\cite{Espeholt:2018} & mins/hours & 200M & 250K & 100-200 cores, 1 GPU &\\
\midrule
  CuLE (\emph{emulation only}) & N/A & N/A & 41K-155K& System I (1 GPU) &\\
  CuLE (\emph{inference only}, A2C, single batch) & N/A & N/A & 39K-125K & System I (1 GPU) &\\
%CuLE (PPO) & hours & <200M & 20K & System II (1 GPU) & Y & 1024 ALEs\\
  CuLE (\emph{training}, A2C~+~V-trace, multiple batches) & 1 hour & 200M & 26K-68K & System I (1 GPU) &\\
  CuLE (\emph{training}, A2C~+~V-trace, multiple batches)* & mins & 200M & 142-187K & System III (4 GPUs) &\\
\bottomrule\\
\multicolumn{6}{l}{*FPS measured on Asterix, Assault, MsPacman, and Pong.}
\end{tabular}}
%\end{sc}
%\end{tiny}
\end{table}

%\begin{wraptable}{R}{0.5\textwidth}
\begin{table}
\caption{Systems used for experiments.}
\label{tab:systems}
\centering
\begin{footnotesize}
\scalebox{0.8}{
\begin{tabular}{ccc}
\toprule
System & Intel CPU & NVIDIA GPU\\
\midrule
I & 12-core Core i7-5930K @3.50GHz & Titan V\\
II & 6-core Core i7-8086K @5GHz & Tesla V100\\
III & 20-core Core E5-2698 v4 @2.20GHz $\times$ 2 & Tesla V100 $\times$ 8, NVLink\\
\bottomrule
\end{tabular}}
\end{footnotesize}
\end{table}
%\end{wraptable}

We focus our attention on the \emph{inference path} and move from the
traditional CPU implementation of the Atari Learning Environment (ALE), a set
of Atari 2600 games that emerged as an excellent DRL
benchmark~\cite{Bellemare:2013,Machado:2017}.  We show that significant
performance bottlenecks primarily stem from CPU environment emulation: the CPU
cannot run a large set of environments simultaneously, the CPU-GPU
communication bandwidth is limited, and the GPU is consequently underutilized.
To both investigate and mitigate these limitations we introduce CuLE (CUDA
Learning Environment), a DRL library containing a CUDA enabled Atari 2600
emulator, that renders frames directly in the GPU memory, avoids off-chip
communication and achieves high GPU utilization by processing thousands of
environments in parallel---something so far achievable only through large and
costly distributed systems.  Compared to the traditional CPU-based approach,
GPU emulation improves the utilization of the computational resources: CuLE on
a single GPU generates more Frames Per Second\footnote{Raw frames are reported
here and in the rest of the paper, unless otherwise specified. These are the
frames that are actually emulated, but only 25\% of them are rendered and used
for training.  Training frames are obtained dividing the raw frames by 4---see
also~\cite{Espeholt:2018}.} (FPS) on the \emph{inference path} (between 39K and
125K, depending on the game, see Table~\ref{tab:compute}) compared to its CPU
counterpart (between 12.5K and 19.8K).  Beyond offering CuLE (\culeurl) as
a tool for research in the DRL field, our contribution can be summarized as
follow:

{\bf(1)} We identify common computational bottlenecks in several DRL
implementations that prevent effective utilization of high throughput compute
units and effective scaling to distributed systems.

{\bf(2)} We introduce an effective batching strategy for large environment
sets, that allows leveraging the high throughput generated by CuLE to quickly
reach convergence with A2C+V-trace~\cite{Espeholt:2018}, and show effective
scaling on multiple GPUs.  This leads to the consumption of 26-68K FPS along
the \emph{training path} on a single GPU, and up to 187K FPS using four GPUs,
comparable (Table~\ref{tab:compute}) to those achieved by large
clusters~\cite{Sto18,Espeholt:2018}.

{\bf(3)} We analyze advantages and limitations of GPU emulation with CuLE in
DRL, including the effect of thread divergence and of the lower (compared to
CPU) number of instructions per second per thread, and hope that our insights
may be of value for the development of efficient DRL systems.

\section{Related Work}~\label{sec:related_work}

The wall clock convergence time of a DRL algorithm is determined by two main
factors: its \emph{sample efficiency}, and the \emph{computational efficiency}
of its implementation.  Here we analyze the sample and computational efficiency
of different DRL algorithms, in connection with their implementation.

We first divide DRL algorithms into policy gradient and Q-value methods, as
in~\cite{Stooke:2018}.  Q-learning optimizes the error on the estimated action
values as a proxy for policy optimization, whereas policy gradient methods
directly learn the relation between a state, $s_t$, and the optimal action,
$a_t$; since  at each update they follow, by definition, the gradient with
respect to the policy itself, they improve the policy more efficiently.  Policy
methods are also considered more general, e.g. they can handle continuous
actions easily.  Also the on- or off-policy nature of an algorithm profoundly
affects both its sample and computational efficiency.  Off-policy methods allow
re-using experiences multiple times, which directly improves the sample
efficiency; additionally, old data stored in the GPU memory
% (replay buffer in Fig.~\ref{fig:design})
can be used to continuously update the DNN on the GPU,
leading to high GPU utilization without saturating the \emph{inference path}.
The replay buffer has a positive effect on the stability of learning as
well~\cite{Mnih:2015}.  On-policy algorithms saturate the \emph{inference path}
more easily, as frames have to be generated on-the-fly using the current policy
and moved from the CPU emulators to the GPU for processing with the DNN.
On-policy updates are generally effective but they are also more prone to fall
into local minima because of noise, especially if the number of environment is
small --- this is the reason why on-policy algorithms largely benefit (in term
of stability) from a significant increase of the number of environments.

Policy gradient algorithms are often on-policy: their efficient update strategy
is counterbalanced by the bottlenecks in the \emph{inference path} and
competition for the use of the GPU along the \emph{inference} and
\emph{training path} at the same time.  Acceleration by scaling to
a distributed system is possible but inefficient in this case: in
IMPALA~\cite{Espeholt:2018} a cluster with hundreds of CPU cores is needed to
accelerate A2C, while training is desynchronized to hide latency.  As
a consequence, the algorithm becomes off-policy, and V-trace was introduced to
deal with off-policy data (see details in the Appendix).  Acceleration on
a DGX-1 has also been demonstrated for A2C and PPO, using large batch sizes to
increase the GPU occupancy, and asynchronous distributed models that hide
latency, but require periodic updates to remain synchronized~\cite{Stooke:2018}
and overall achieves sublinear scaling with the number of GPUs.
\section{CUDA Learning Environment (CuLE)}
~\label{sec:cule}

\begin{wrapfigure}{R}{0.55\textwidth}
% \begin{figure}
\centering
\includegraphics[trim=0.25cm 10.5cm 0.1cm 0.8cm,clip, width=0.49\textwidth]{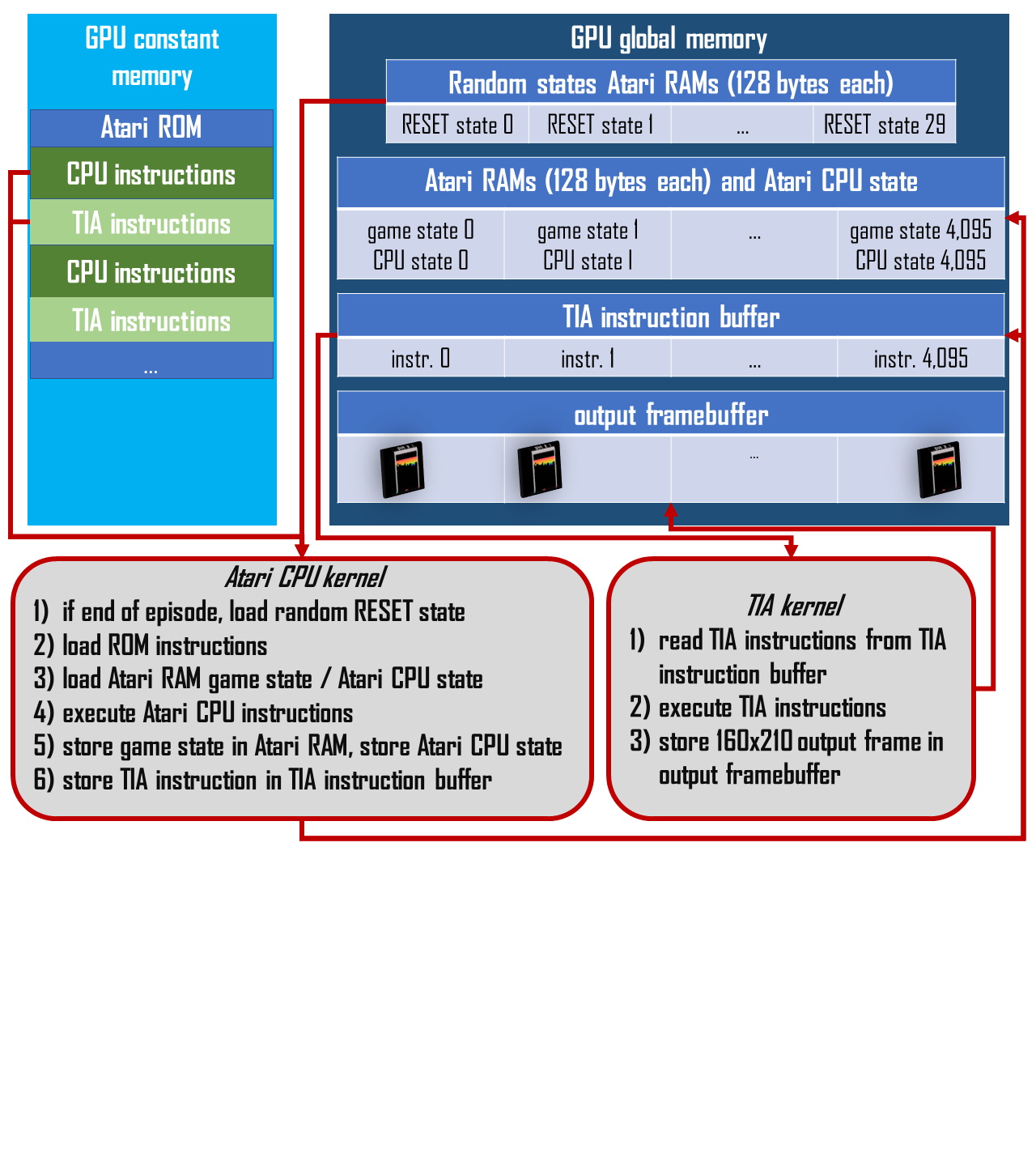}
\vspace{-0.3cm}

\caption{Our CUDA-based Atari emulator uses an \emph{Atari CPU kernel} to
  emulate the functioning of the Atari CPU and advance the game state, and
  a second \emph{TIA kernel} to emulate the TIA and render frames directly in
  GPU memory. For episode resetting we generate and store a cache of random
  initial states. Massive parallelization on GPU threads allows the parallel
  emulation of thousands of Atari games.}

\label{fig:cule_overview}
% \end{figure}
\end{wrapfigure}

In CuLE, we emulate the functioning of many Atari consoles in parallel using
the CUDA programming model, where a sequential host program executes parallel
programs, known as kernels, on a GPU.  In a trivial mapping of the Atari
emulator to CUDA, a single thread emulates both the Atari CPU and TIA to
execute the ROM code, update the Atari CPU and TIA registers as well as the
game state in the 128 bytes RAM, and eventually render the pixels in the output
frame.  However, the contrasting nature of the game code execution and
renderings tasks, the first dominated by reading from the RAM/ROM and writing
tens of bytes to RAM, while the second writes hundreds of pixels to the
framebuffer, poses a serious issue in terms of performance, such as thread
divergence and an imbalanced number of registers required by the first and
second tasks.  To mitigate these issues, CuLE uses two CUDA kernels: the first
one first loads data from the GPU global memory, where we store the state of
each emulated Atari processor, and the 128 bytes RAM data containing the
current state of the game; it also reads ROM instructions from the constant
GPU memory, executes them to update the Atari CPU and game states, and stores
the updated Atari CPU and game states back into GPU global memory. It is
important to notice that this first kernel does not execute the TIA
instructions read from the ROM, but copies them into the  TIA instruction
buffer in GPU global memory, which we implemented to decouple the execution of
the Atari CPU and TIA instructions in CuLE.  The second CuLE kernel emulates
the functioning of the TIA  processor: it first reads the instructions stored
in the TIA instruction buffer, execute them to update the TIA registers, and
renders the $160 \times 210$ output framebuffer in global GPU memory.  Despite
this implementation requires going through the TIA instruction twice, it has
several advantages over the single-kernel trivial implementation.  First of
all, the requirements in terms of registers per thread and the chance of having
divergent code are different for the Atari CPU and TIA kernels, and the use of
different kernels achieves a better GPU usage.  A second advantage that we
exploit is that not all frames are rendered in ALE: the input of the RL
algorithm is the pixelwise maximum between the last two frames in a sequence of
four, so we can avoid calling the TIA kernel when rendering of the screen is
not needed.  A last advantage, not exploited in our implementation yet, is that
the TIA kernel may be scheduled one the GPU with more than one thread per game,
as rendering of diverse rows on the screen is indeed a parallel operation - we
leave this optimization for future developments of CuLE. 

%all the data related to the processor, ROM, and RAM, and  executes all instructions
%sequentially to update the game state data with the exception of the
%framebuffer.  Instead of writing updates to the framebuffer, each thread writes
%in a global buffer the updates it would have made to Television Interface
%Adaptor (TIA).  The TIA is a secondary processor embedded in the Atari emulator
%whose aim is to translate these updates into frames on the display---we emulate
%it through a second CUDA kernel, that reads from the cached buffer and generate
%the frames.

To better fit our execution model, our game reset strategy is also different
from the one in the existing CPU emulators, where 64 startup frames are
executed at the end of each episode.  Furthermore, wrapper interfaces for RL,
such as ALE, randomly execute an additional number of frames (up to 30) to
introduce randomness into the initial state.  This results into up to 94 frames
to reset a game, which may cause massive divergence between thousands of
emulators executing in SIMD fashion on a GPU.  To address this issue, we
generate and store a cache of random initial states (30 by default) when a set
of environments is initialized in CuLE.  At the end of an episode, each
emulator randomly selects one of the cached states as a seed and copies it into
the terminal emulator state.

Some of the choices made for the implementation of CuLE are informed by ease of
debugging, like associating one state update kernel to one environment, or need
for flexibility, like emulating the Atari console instead of directly writing
CUDA code for each Atari game.  A 1-to-1 mapping between threads and emulators
is not the most computationally efficient way to run Atari games on a GPU, but
it makes the implementation relatively straightforward and has the additional
advantage that the same emulator code can be executed on the CPU for debugging
and benchmarking (in the following, we will refer to this implementation as
CuLE\textsubscript {CPU}).  Despite of this, the computational advantage
provided by CuLE over traditional CPU emulation remains significant.
%Specific CUDA implementations of any environment may be even more
%computationally efficient, although less flexible. This remains an interesting
%future direction of research.

\section{Experiments}~\label{sec:experiments}

\paragraph{Atari emulation} We measure the FPS under different conditions: we
get an upper bound on the maximum achievable FPS in the \emph{emulation only}
case, when we emulate the environments and use a random policy to select
actions.  In the \emph{inference only} case, we measure the FPS along the
\emph{inference path}: a policy DNN selects the actions, CPU-GPU data transfer
occur for CPU emulators, while both emulation and DNN inference run on the GPU
when CuLE is used.  This is the maximum throughput achievable by off-policy
algorithms, when data generation and consumption are decoupled and run on
different devices.  In the \emph{training} case, the entire DRL system is at
work: emulation, inference, and training may all run on the same GPU.  This is
representative of the case of on-policy algorithms, but the FPS are also
affected by the computational cost of the specific DRL update algorithm; in our
experiments we use a vanilla A2C~\cite{openaiblog}, with N-step bootstrapping,
and $N=5$ as the baseline (for details of A2C and off-policy correction with
V-trace, see the Appendix).
%---we analyze in detail this complex case in the last paragraph of this
%section, which also shows how to leverage CuLE at best to fully utilize one or
%more GPUs and reach convergence in a short time (see
%Table~\ref{tab:a2c_vtrace}).  Table~\ref{tab:systems} shows the systems used
%in our experiments.

\begin{figure*}[!ht]
\centering
%\captionsetup{position=top}
\begin{tabular}{cc}
FPS & FPS per environment\\ 
\begin{subfigure}[\emph{emulation only}]
{\includegraphics[width=0.23\textwidth]{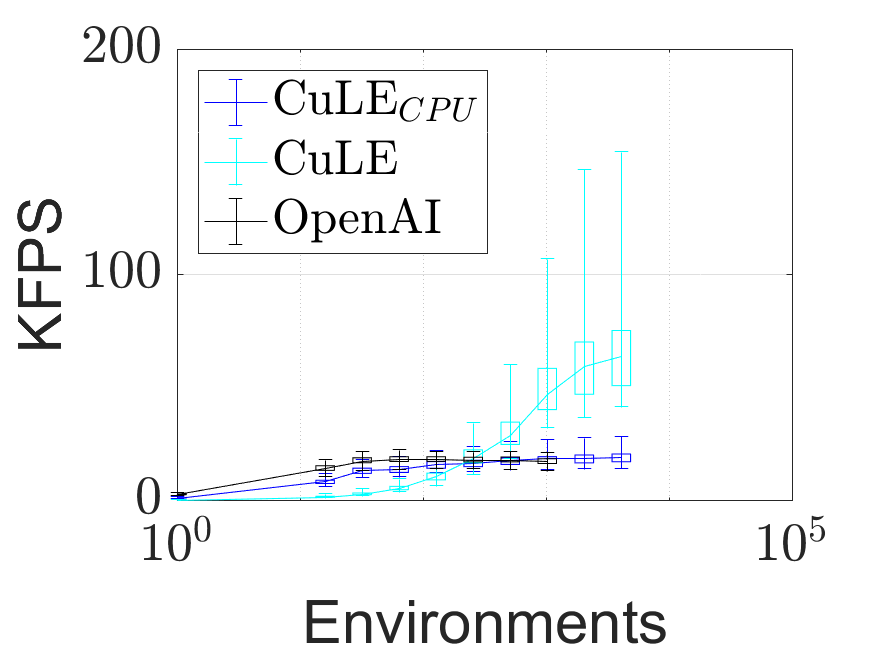}
\label{fig:raw_fps_random}}
\end{subfigure}
\begin{subfigure}[\emph{inference only}]{\includegraphics[width=0.23\textwidth]{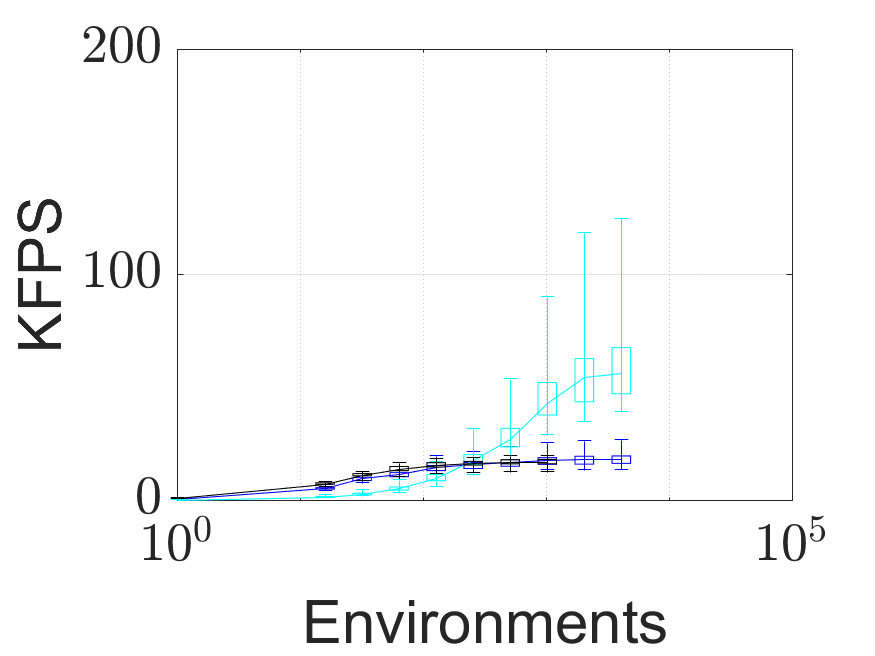}\label{fig:raw_fps_inference}}
\end{subfigure}
&\begin{subfigure}
[\emph{emulation only}]{\includegraphics[width=0.23\textwidth]{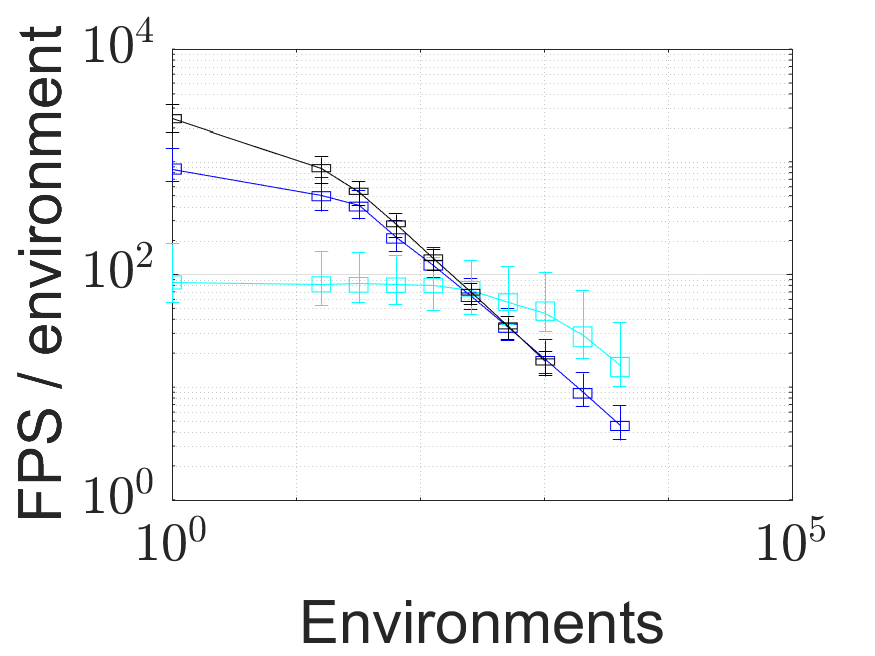}
\label{fig:raw_fps_per_env_random}}
\end{subfigure}
\begin{subfigure}[\emph{inference only}]{\includegraphics[width=0.23\textwidth]{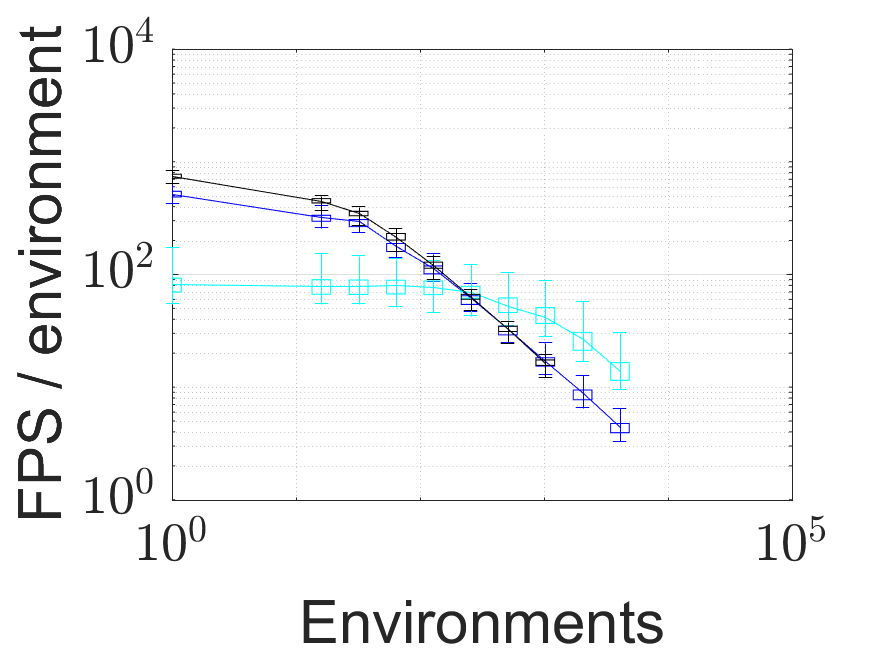}\label{fig:raw_fps_per_env_inference}}
\end{subfigure}
\end{tabular}
\caption{FPS and FPS / environment on System I in Table~\ref{tab:systems},
  for OpenAI Gym~\cite{openaiblog}, CuLE\textsubscript{CPU}, and CuLE, as a function of the number of environments, under  different load conditions: \emph{emulation only}, and \emph{inference only}. The boxplots indicate the minimum, $25^{th}$, $50^{th}$, $75^{th}$ percentiles and maximum FPS, for the entire set of 57 Atari games.}
\label{fig:raw_fps}
\end{figure*}

\begin{table*}

\caption{Training FPS, DNN's Update Per Second (UPS), time to reach a given
  score, and corresponding number of training frames for four Atari games,
  A2C+V-trace, and different configurations of the emulation engines, measured
  on System I in Table~\ref{tab:systems} (System III for the multi-GPU case).
  The best metric in each row is in bold.}

\label{tab:a2c_vtrace}
\centering
\begin{scriptsize}
\begin{tabular}{r|cccc|ccc|c|c}
\toprule
Engine & \multicolumn{4}{c|}{OpenAI Gym} & \multicolumn{3}{c|}{CuLE, 1 GPU} & CuLE, 4 GPUs & Game \\
\midrule
Envs & 120 & 120 & 120 & 1200 & 1200 & 1200 & 1200 & 1200$\times$4 & \multirow{4}{*}{\rotatebox[origin=c]{90}{---}} \\
Batches & 1 & 5 & 20 & 20 & 1 & 5 & 20 & 20$\times$4 &\\
N-steps & 5 & 5 & 20 & 20 & 5 & 5 & 20 & 20 &\\
SPU & 5 & 1 & 1 & 1 & 5 & 1 & 1 & 1 &\\
\midrule
Training KFPS & 4.2 & 3.4 & 3.0 & 4.9 & 10.6 & 11.5 & 11.0 & \bf{42.7} & \multirow{4}{*}{\rotatebox[origin=c]{90}{Assault}}\\
UPS & 7.0 & \bf{28.3} & 24.7 & 4.1 & 1.8 & 9.6 & 9.1 & 8.9 \\
Time [mins] & 20.2 & --- & 42.6 & 44.2 & 18.8 & 9.4 & 9.9 & \bf{7.9} \\
Training Mframes (for average score: 800) & \bf{5.0} & --- & 7.5 & 13.0 & 12.0 & 6.5 & 6.5 & 18.0 \\
\midrule
Training KFPS & 4.3 & 3.3 & 3.0 & 4.9 & 11.9 & 12.5 & 12.1 & \bf{46.6} & \multirow{4}{*}{\rotatebox[origin=c]{90}{Asterix}}\\
UPS & 7.1 & \bf{27.9} & 24.8 & 4.1 & 2.0 & 10.4 & 10.0 & 9.7 \\
Time [mins] & 8.1 & 35.2 & 14.4 & 27.1 & --- & 14.0 & 3.4 & \bf{2.5} \\
Training Mframes (for average score: 1,000) & \bf{2.0} & 7.0 & 2.5 & 8.0 & --- & 10.5 & 2.5 & 7.0 \\
\midrule
Training KFPS & 4.0 & 3.3 & 2.8 & 4.8 & 9.0 & 9.6 & 9.2 & \bf{35.5} & \multirow{4}{*}{\rotatebox[origin=c]{90}{MsPacman}}\\
UPS & 6.7 & \bf{27.1} & 23.7 & 4.0 & 1.5 & 8.0 & 7.7 & 7.4 \\
Time [mins] & 16.6 & 20.5 & 14.7 & 12.4 & --- & 6.9 & 11.8 & \bf{2.4} \\
Training Mframes (for average score: 1,500) & 4.0 & 4.0 & \bf{2.5} & 3.5 & --- & 4.0 & 6.5 & 3.0 \\
\midrule
Training KFPS & 4.3 & 3.4 & 3.0 & 4.8 & 10.5 & 11.2 & 10.6 & \bf{41.7K} & \multirow{4}{*}{\rotatebox[origin=c]{90}{Pong}}\\
UPS & 7.2 & \bf{28.1} & 24.9 & 4.0 & 1.8 & 9.3 & 8.9 & 8.7 \\
Time [mins] & 21.2 & 12.2 & 8.4 & 8.7 & --- & 5.9 & 3.1 & \bf{2.4} \\
Training Mframes (for average score: 18) & 5.5 & 2.5 & \bf{1.5} & 2.5 & --- & 4.0 & 2.0 & 6.0 \\
\bottomrule
\end{tabular}
\end{scriptsize}
\end{table*}

Figs.~\ref{fig:raw_fps_random}-\ref{fig:raw_fps_inference} show the FPS
generated by OpenAI Gym, CuLE\textsubscript{CPU}, and CuLE, on the entire set of Atari
games, as a function of the number of environments.  In the \emph{emulation
only} case, CPU emulation is more efficient for a number of environments up to
128, when the GPU computational power is not leveraged because of the low
occupancy.  For a larger number of environments, CuLE significantly overcomes
OpenAI Gym, for which FPS are mostly stable for 64 environments or more,
indicating that the CPU is saturated: the ratio between the
median FPS generated by CuLE with 4096 environment (64K) and the peak FPS for
OpenAI Gym (18K) is $3.56\times$.  In the \emph{inference only} case there are
two additional overheads: CPU-GPU communication (to transfer observations), and
DNN inference on the GPU. Consequently, CPU emulators achieve a lower FPS in
\emph{inference only} when compared to \emph{emulation only}; the effects of
the overheads is more evident for a small number of environments, while the FPS
slightly increase with the number of environments without reaching the
\emph{emulation only} FPS.  CuLE's FPS are also lower for \emph{inference
only}, because of the latency introduced by DNN inference, but the FPS grow
with the number of environments, suggesting that the computational capability
of the GPU is still far from being saturated.

~\begin{figure*}[!ht]
\centering
  \subfigure[Asterix, GPU]{\includegraphics[trim=0.3cm 0cm 1.2cm 0.3cm,clip, width=0.235\textwidth]{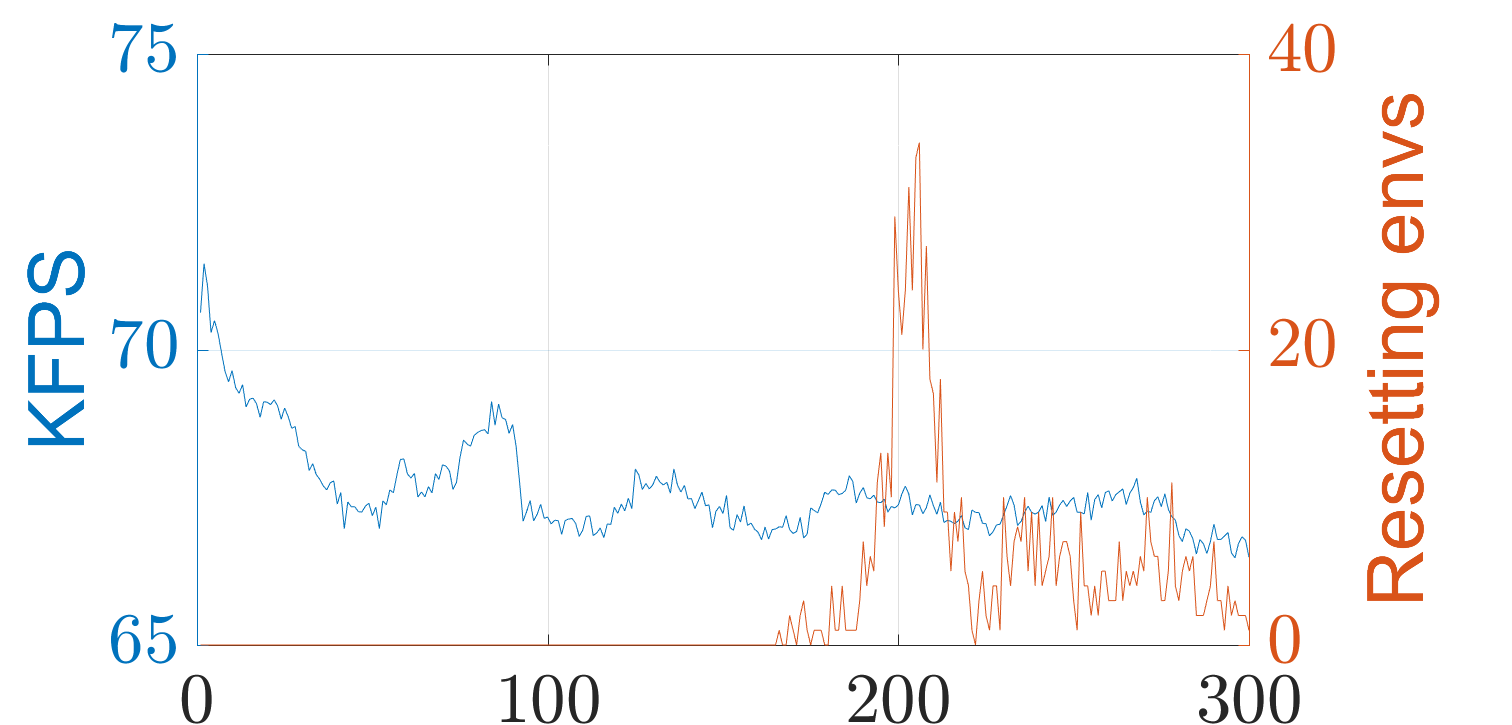}~\label{fig:decor_asterix_gpu}}
  \subfigure[Pong, GPU]{\includegraphics[trim=0.3cm 0cm 1.2cm 0.3cm,clip,width=0.235\textwidth]{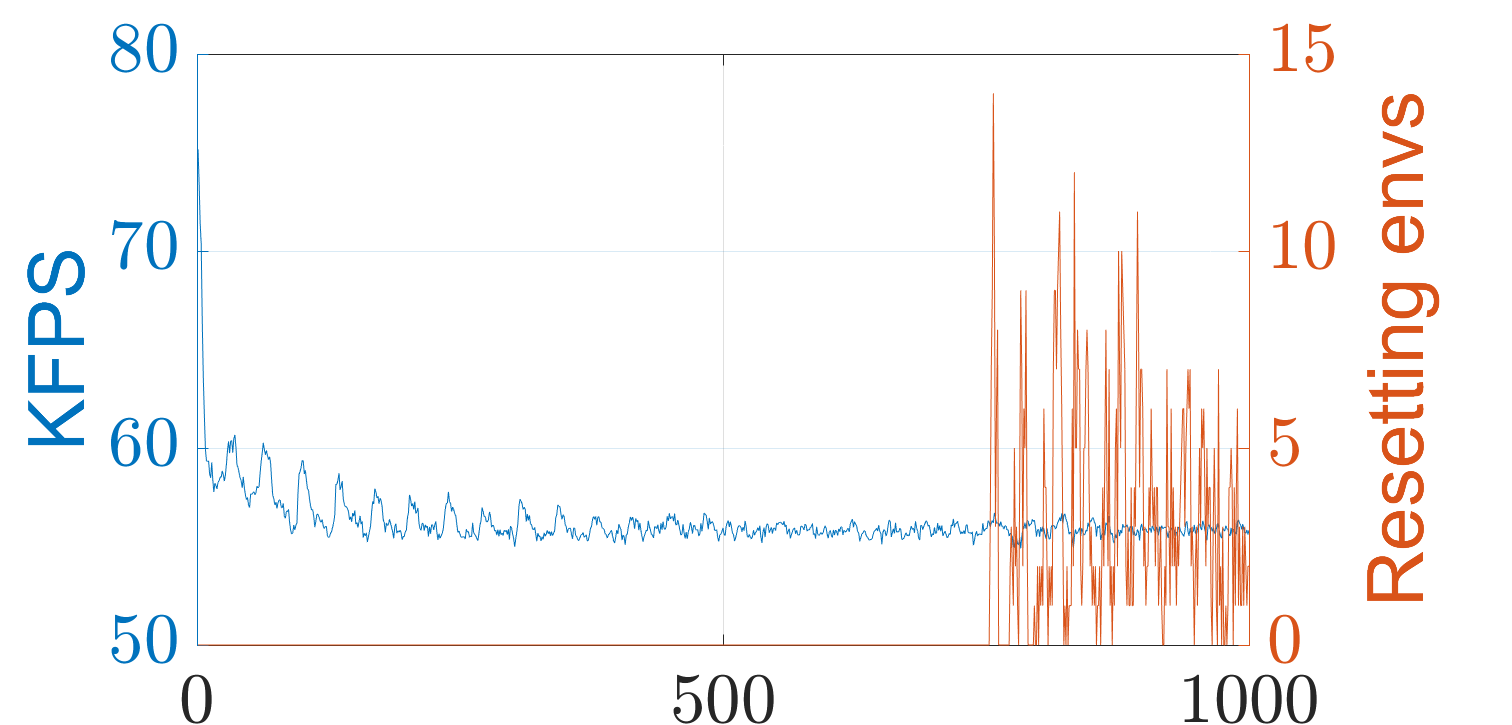}~\label{fig:decor_pong_gpu}}
  \subfigure[Ms Pacman, GPU]{\includegraphics[trim=0.3cm 0cm 1.2cm 0.3cm,clip,width=0.235\textwidth]{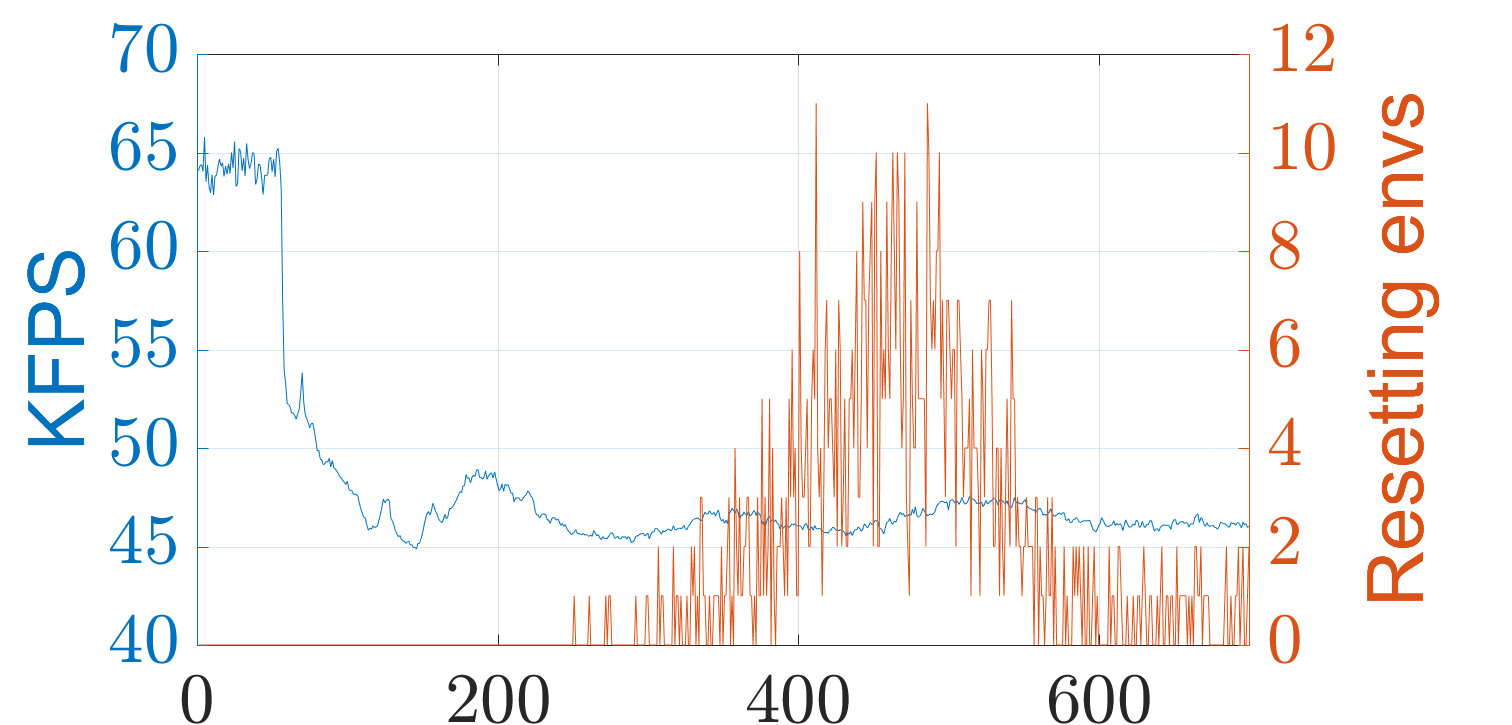}~\label{fig:decor_mspacman_gpu}}
  \subfigure[Assault, GPU]{\includegraphics[trim=0.3cm 0cm 1.2cm 0.3cm,clip,width=0.235\textwidth]{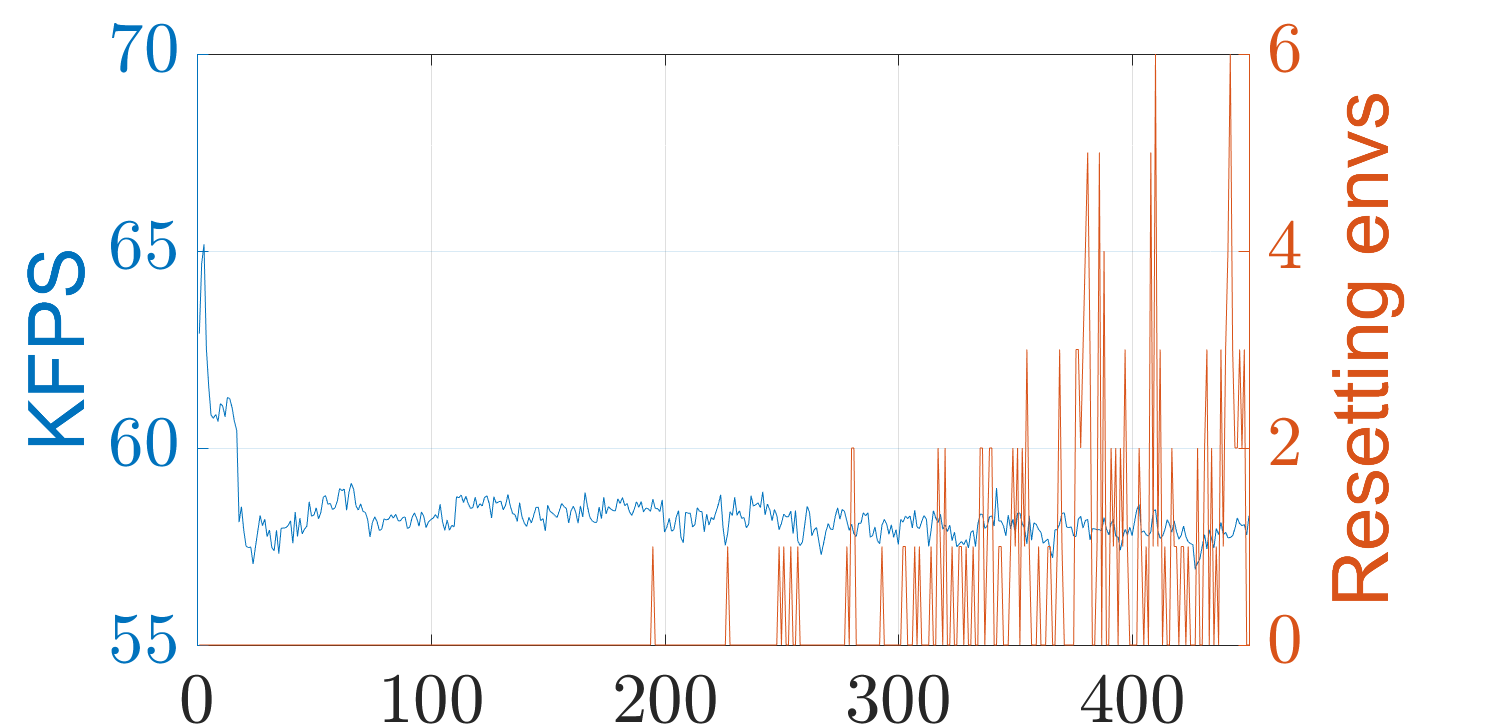}~\label{fig:decor_assault_gpu}}
%  \\
%  \subfigure[Asterix, CPU]{\includegraphics[width=0.23\textwidth]{figures/decorrelation/Decorrelation_Asterix_CPU.png}~\label{fig:decor_asterix_cpu}}
%  \subfigure[Pong, CPU]{\includegraphics[width=0.23\textwidth]{figures/decorrelation/Decorrelation_Pong_CPU.png}~\label{fig:decor_pong_cpu}}
%  \subfigure[Ms Pacman, CPU]{\includegraphics[width=0.23\textwidth]{figures/decorrelation/Decorrelation_MsPacman_CPU.png}~\label{fig:decor_mspacman_cpu}}
%  \subfigure[Assault, CPU]{\includegraphics[width=0.23\textwidth]{figures/decorrelation/Decorrelation_Assault_CPU.png}~\label{fig:decor_assault_cpu}}
%

\caption{FPS as a function of the environment step, measured on System I in
  Table~\ref{tab:systems} for \emph{emulation only} on four Atari games, 512
  environments, for CuLE; each panel also shows the number of resetting
  environments.  FPS is higher at the beginning, when all environments are in
  similar states and thread divergence within warps is minimized; after some
  steps, correlation is lost, FPS decreases and stabilizes. Minor oscillations
  in FPS are possibly associated to more or less computational demanding phases
  in the emulation of the environments (e.g.,  when a goal is scored in Pong).}

\label{fig:decor}
\end{figure*}

\paragraph{Factors affecting the FPS}
Figs.~\ref{fig:raw_fps_random}-\ref{fig:raw_fps_inference} shows that the
throughput varies dramatically across games: 4096 CuLE\textsubscript{CPU}
environments run at 27K FPS on Riverraid, but only 14K FPS for Boxing:
a $1.93\times$ difference, explained by the different complexity of the ROM
code of each game.  The ratio between the maximum and minimum FPS is amplified
in the case of GPU emulation: Riverraid runs in \emph{emulation only} at 155K
FPS when emulated by CuLE and 4096 environments, while UpNDown runs at 41K FPS
---a $3.78\times$ ratio.
% We believe that thread divergence on the GPU plays an important role in this:
% in fact, the GPU's SIMT (Single Instruction Multiple Thread) architecture
% serializes the execution of threads in the same warp along different code
% branches, decreasing the total instruction throughput.  In practice, beyond
% code complexity which is in common with CPU emulation, a different branching
% factor among games amplifies their difference in terms FPS. 

To better highlight the impact of thread divergence on throughput, we measure
the FPS for CuLE, \emph{emulation only}, 512 environments, and four games
(Fig.~\ref{fig:decor}). All the environments share the same initial state, but
random action selection leads them to diverge after some steps.  Each
environment resets at the end of an episode.  The FPS is maximum at the very
beginning, when all the environments are in similar states and the chance to
execute the same instruction in all the threads is high.  When they move
towards different states, code divergence negatively impacts the FPS, until it
reaches an asymptotic value.  This effect is present in all games and
particularly evident for MsPacman in Fig.~\ref{fig:decor}; it is not present in
CPU emulation (see Appendix).  Although divergence can reduce FPS by
$30\%$ in the worst case, this has to be compared with case of complete
divergence within each thread and for each instruction, which would yield $1/32
\simeq 3\%$ of the peak performances.  Minor oscillations of the FPS are also
visible especially for games with a repetitive pattern (e.g. Pong), where
different environments can be more or less correlated with a typical
oscillation frequency.

\begin{wrapfigure}{R}{0.45\textwidth}
% \begin{figure}
\centering
  \subfigure{\includegraphics[trim=3cm 0.2cm 3cm 1.2cm, clip, width=0.4\textwidth]{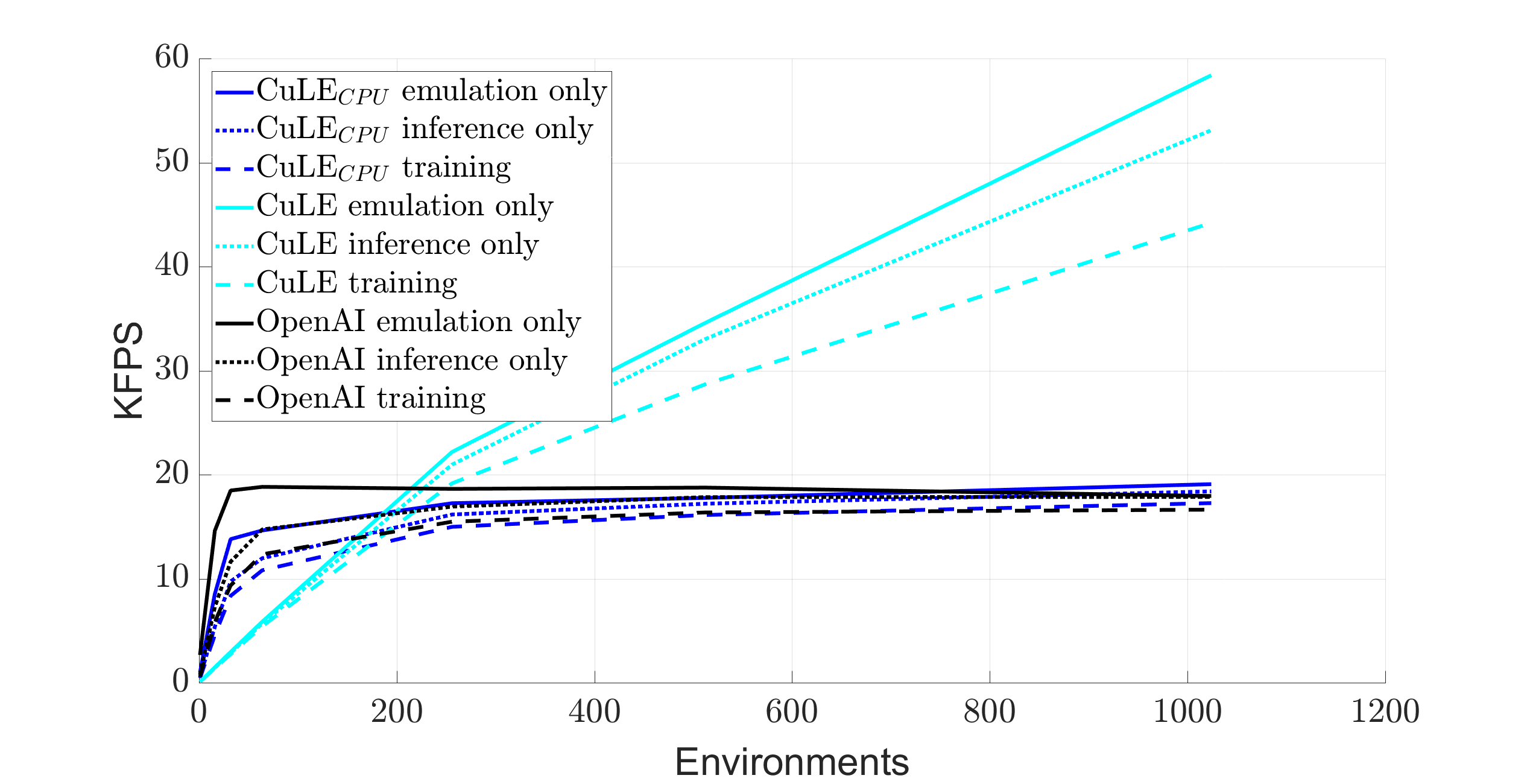}~\label{fig:bench_assault}}
\caption{FPS generated by different emulation engines on System I in
  Table~\ref{tab:systems} for Assault, as a function of the
  number of environments, and different load conditions for A2C with N-step bootstrapping, $N=5$).}
  \label{fig:bench_training}
% \end{figure}
\end{wrapfigure}

\paragraph{Performances during training}

Fig.~\ref{fig:bench_training} compares the FPS generated by different emulation
engines on a specific game (Assault)\footnote{Other games for which we observe
a similar behavior are reported  in the Appendix, for sake of space.}, for
different load conditions, including the \emph{training} case, and number of
environments.  As expected, when the entire \emph{training path} is at work,
the FPS decreases even further.  However, for CPU emulators, the difference
between FPS in the \emph{inference only} and \emph{training} cases decreases
when the number of environments increases, as the system is bounded by the CPU
computational capability and CPU-GPU communication bandwidth.  In the case of
the CPU scaling to multiple GPUs would be ineffective for on-policy algorithms,
such GA3C~\cite{Babaeizadeh:2016,Babaeizadeh:2017}, or sub-optimal, in the case
of distributed systems~\cite{Espeholt:2018, Stooke:2018}.  On the other hand,
the difference between \emph{inference only} and \emph{training} FPS increases
with the number of environments for CuLE, because of the additional training
overhead on the GPU.  The potential speed-up provided by CuLE for vanilla A2C
and Assault in Fig.~\ref{fig:bench_training} is $2.53\times$ for 1,024
environments, but the system is bounded by the GPU computational power; as
a consequence, better batching strategies that reduce the training
computational overhead as well as scaling to multiple GPUs are effective to
further increase the speed-up ratio, as demonstrated later in this Section.

When data generation and training can be decoupled, like for off-policy
algorithms, training can be easily moved to a different GPU and the
\emph{inference path} can be used at maximum speed.  The potential speed-up
provided by CuLE for off-policy algorithms is then given by the ratio between
the \emph{inference only} median FPS for CuLE (56K) and CuLE\textsubscript{CPU} (18K),
which is $3.11\times$ for 4,096 environments.  Furthermore, since the FPS
remains flat for CPU emulation, the advantage of CuLE amplifies (for both on-
and off-policy methods) with the number of environments.

%Table~\ref{tab:compute} shows that the average FPS generated by CuLE during
%\emph{emulation only} and \emph{inference only} on a single GPU is comparable
%to that achieved by large, and more costly, distributed systems.  The same
%Table reports the FPS for CuLE during training with different DRL algorithms
%- in the case of PPO on a single GPU, our implementation suffers from an
%additional synchronization overhead (due to the on-policy nature of the
%algorithm that put the data generation and training engines in competition for
%the GPU computational resources) that does not affect others. Off-policy
%algorithms implemented on distributed systems are reported in the same Table.

\paragraph{Frames per second per environment}

Fig.~\ref{fig:raw_fps_per_env_random}-\ref{fig:raw_fps_per_env_inference} show
the FPS / environment for different emulation engines on System I, as
a function of the number of  environments.  For 128 environments or fewer, CPU
emulators generate frames at a higher rate (compared to CuLE), because CPUs are
optimized for low latency, and execute a high number of instructions per second
per thread.  However, the FPS / environment decrease with the number of
environments, that have to share the same CPU cores.  Instead, the GPU
architecture maximizes the throughput and has a lower number of instructions
per second per thread.  As a consequence, the FPS / environment is smaller
(compared to CPU emulation) for a small number of environments, but they are
almost constant up to 512 environments, and starts decreasing only after this
point.  In practice, CuLE environments provide an efficient means of training
with a diverse set of data and collect large statistics about the rewards
experienced by numerous agents, and consequently lowering the variance of the
value estimate.  On the other hand, samples are collected less efficiently in
the temporal domain, which may worsen the bias on the estimate of the value
function by preventing the use of large N in N-step bootstrapping.  The last
paragraph of this Section shows how to leverage the high throughput generated
by CuLE, considering these peculiarities.

\paragraph{Memory limitations}

Emulating a massively large number of environments can be problematic
considering the relatively small amount of GPU DRAM.  Our
PyTorch~\cite{PyTorch:2017} implementation of A2C requires each environment to
store 4 84x84 frames, plus some additional variables for the emulator state.
For 16K environments this translates into 1GB of memory, but the primary issue
is the combined memory pressure to store the DNN with 4M parameters and the
meta-data during training, including the past states: training with
16K environments easily exhausts the DRAM on a single GPU (while training on
multiple GPUs increases the amount of available RAM).  Since we did
not implement any data compression scheme as in~\cite{Horgan:2018}, we
constrain our training configuration to fewer than 5K environments, but peak
performance in terms of FPS would be achieved for a higher number of
environments - this is left as a possible future improvement.

\begin{figure*}
\centering
  \subfigure[Assault, 20M training frames]
  {\includegraphics[width=0.33\textwidth]{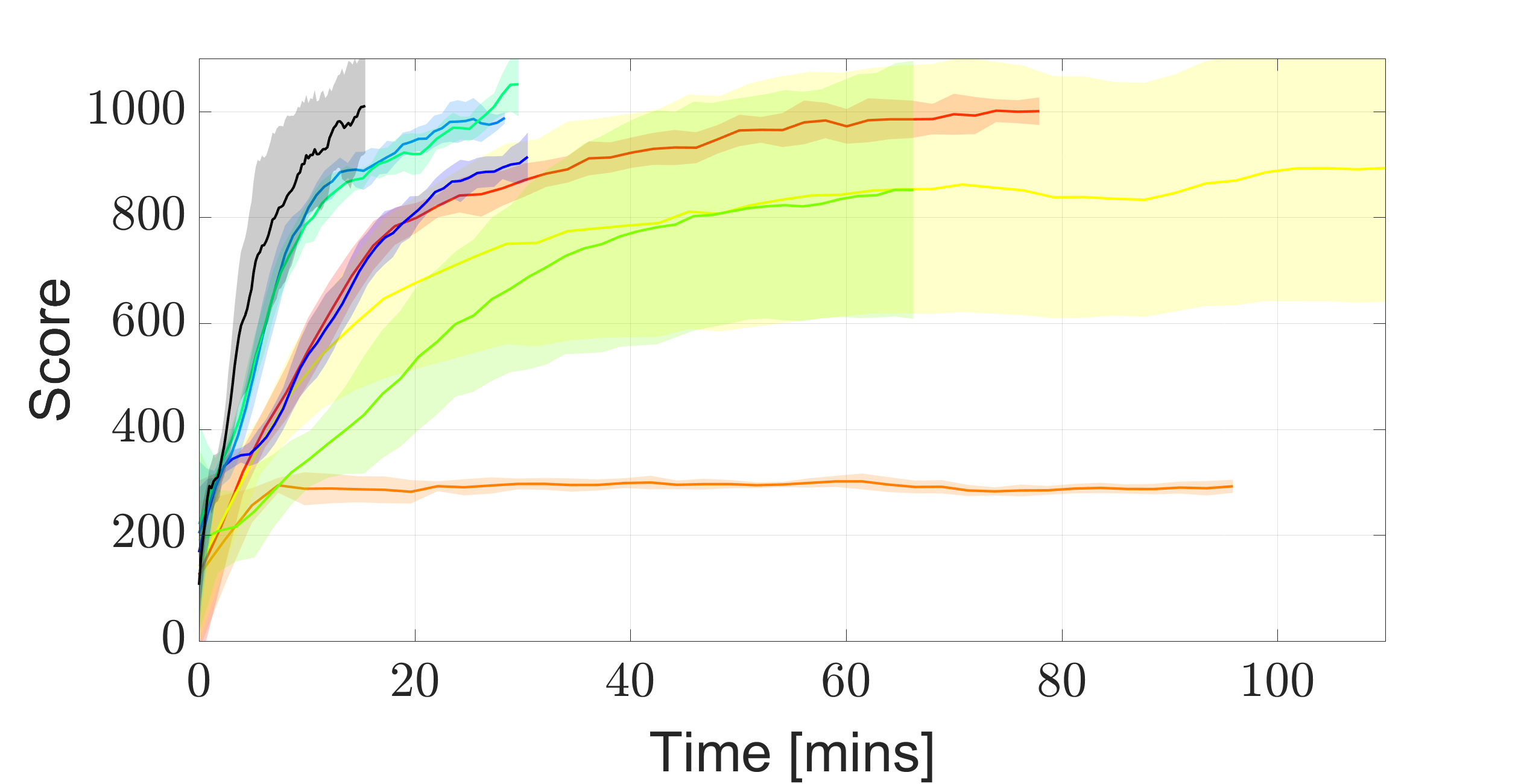}~\label{fig:vtrace_assault}}
  \subfigure[Asterix, 20M training frames]
  {\includegraphics[width=0.33\textwidth]{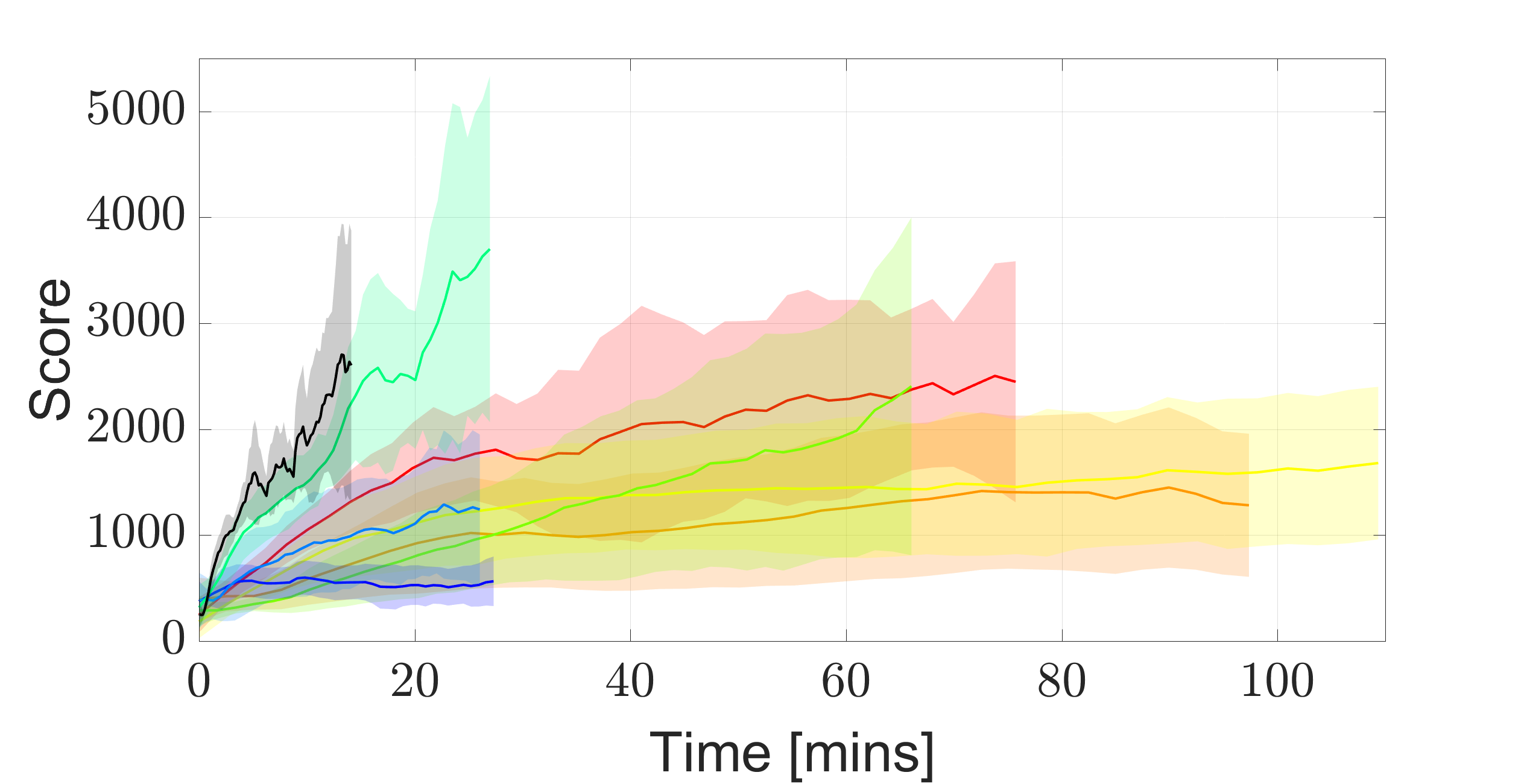}~\label{fig:vtrace_asterix}}
  \subfigure[Ms-Pacman, 20M training frames]
  {\includegraphics[width=0.33\textwidth]{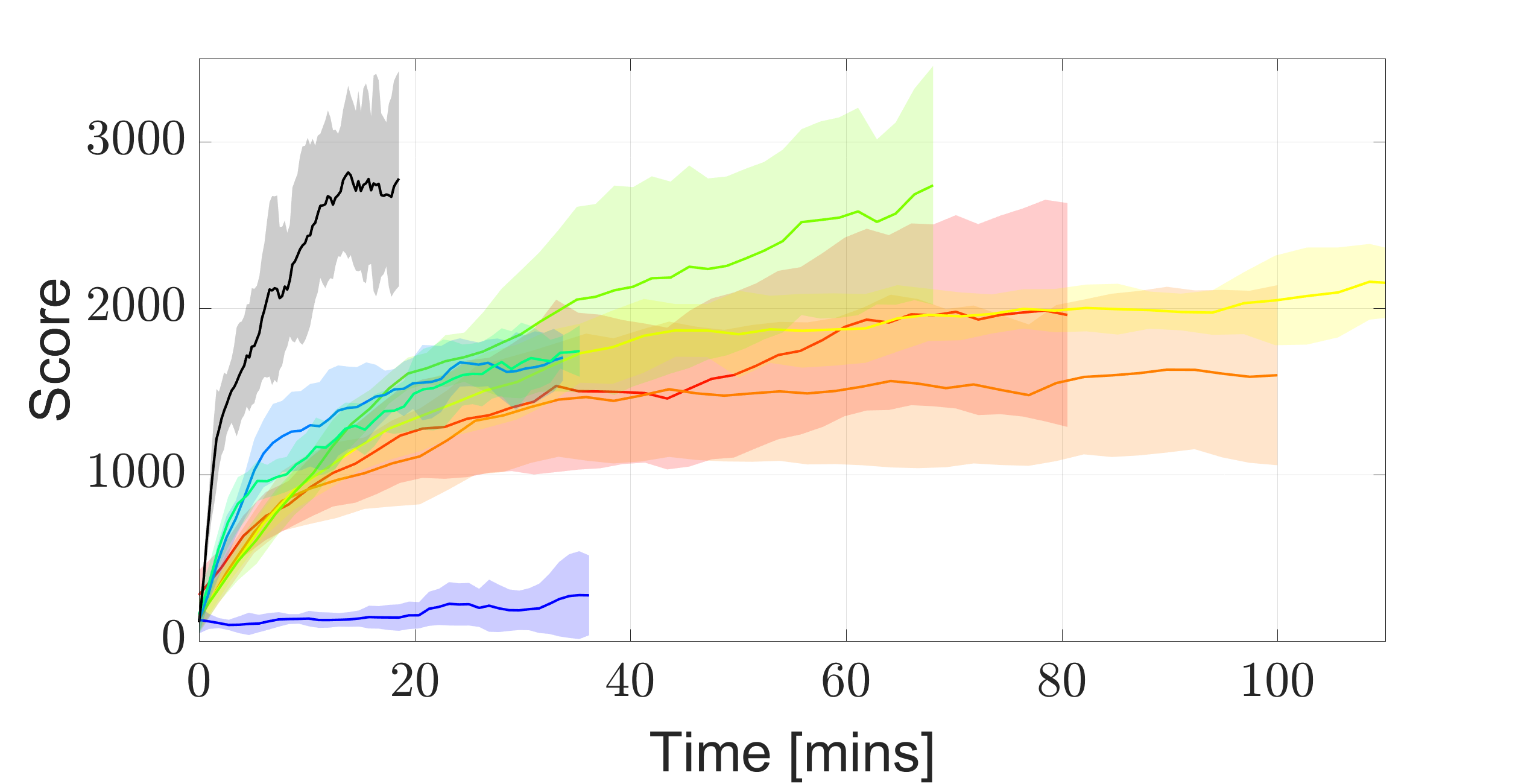}~\label{fig:vtrace_pacman}}
 \subfigure[Pong, 8M training frames]{\includegraphics[width=0.33\textwidth]{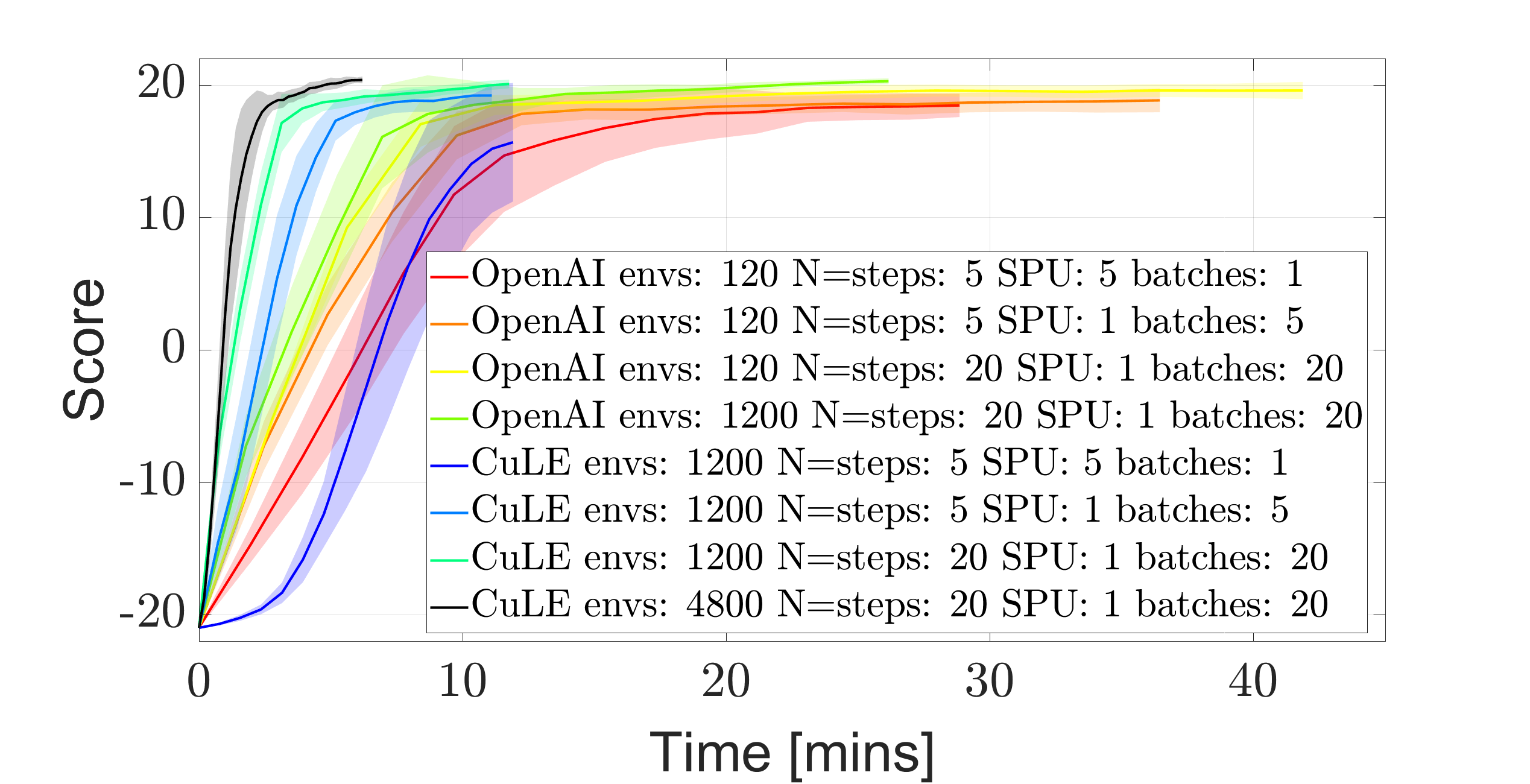}\label{fig:vtrace_pong}}
\caption{Average testing score and standard deviation on four Atari games as a function of the training time, for A2C+V-trace, System III in Table~\ref{tab:systems}, and different batching strategies (see also Table~\ref{tab:a2c_vtrace}). Training frames are double for the multi-GPU case (black line). Training performed on CuLE or OpenAI Gym; testing performed on OpenAI Gym environments (see the last paragraph of Section~\ref{sec:experiments}).}
\label{fig:vtrace}
\end{figure*}

\paragraph{A2C} We analyze in detail the case of A2C with CuLE on a single GPU.
%Beyond showing that faster convergence can be achieved with CuLE, this
%analysis further highlights the interconnections between the computational
%aspects and the convergence properties of the DRL algorithm, thus
%demonstrating how to use CuLE to gain further insights in the DRL field.
As a baseline, we consider vanilla A2C, using 120 OpenAI Gym CPU environments
that send training data to the GPU to update the DNN
% (Fig.~\ref{fig:vtrace_single_batch})
every $N=5$ steps.  This configuration
takes, on average, 21.2 minutes (and 5.5M training frames) to reach a score of
18 for Pong and 16.6 minutes (4.0M training frames) for a score of 1,500 on
Ms-Pacman (Fig.~\ref{fig:vtrace}, red line; first column of
Table~\ref{tab:a2c_vtrace}).  CuLE with 1,200 environments generates
approximately $2.5\times$ more FPS compared to OpenAI Gym, but this alone is
not sufficient to improve the convergence speed (blue line,
Fig.~\ref{fig:vtrace}).  CuLE generates larger batches but, because FPS
/ environment is lower when compared to CPU emulation, fewer Updates Per Second
(UPS) are performed for training the DNN (Table~\ref{tab:a2c_vtrace}), which is
detrimental for learning.

%\begin{wrapfigure}{R}{0.5\textwidth}
% \begin{figure}
% \label{fig:batch_generation}
% \setcounter{subfigure}{0}
% \centering
%   \subfigure[Single batch strategy]{\includegraphics[width=0.4\textwidth]{figures/vtrace/vtrace/Slide1.PNG}\label{fig:vtrace_single_batch}}
%   \subfigure[Multi batch strategy]{\includegraphics[width=0.4\textwidth]{figures/vtrace/vtrace/Slide2.PNG}\label{fig:vtrace_multi_batch}}
% \caption{Different batching strategies  are defined by the number of batches, N-Steps and Steps Per Update (SPU) parameters. The on-policy single batch case (panel a) is a special case of the
% more general, off-policy multi batch approach (panel b). The batching strategy affects both the computational and convergence aspects of the DRL algorithm, as shown in Fig.~\ref{fig:vtrace} and in Table~\ref{tab:a2c_vtrace}.}
% \end{figure}
%\end{wrapfigure}

% ~\begin{wrapfigure}{L}{0.68\textwidth}
% \end{wrapfigure}

\paragraph{A2C+V-trace and batching strategy} To better leverage CuLE, and
similar in spirit to the approach in IMPALA~\cite{Espeholt:2018}, we employ
a different batching strategy
% , illustrated in Fig.~\ref{fig:vtrace_multi_batch}:
% environment steps occur in parallel
on the GPU, but training data are read in batches to update the DNN every Steps Per
Update (SPU) steps.  This batching strategy significantly increases the DNN's
UPS at the cost of a slight decrease in FPS (second columns of OpenAI Gym and
CuLE in Table~\ref{tab:a2c_vtrace}), due to the fact that the GPU has to
dedicate more time to training.  Furthermore, as only the most recent data in
a batch are generated with the current policy, we use
V-trace~\cite{Espeholt:2018} for off-policy correction.  The net result is an
increase of the overall training time when 120 OpenAI Gym CPU environments are
used, as this configuration pays for the increased training and communication
overhead, while the smaller batch size increases the variance in the estimate
of the value function and leads to noisy DNN updates (second column in
Table~\ref{tab:a2c_vtrace}, orange lines in Fig.~\ref{fig:vtrace}).  Since CuLE
does not suffer from the same computational bottlenecks, and at the same time
benefits from the variance reduction associated with the large number (1,200)
of environments, using the same batching strategy with CuLE reduces the time to
reach a score of 18 for Pong and 1,500 for Pacman respectively to 5.9 and 6.9
minutes.  The number of frames required to reach the same score is sometimes
higher for CuLE (Table~\ref{tab:a2c_vtrace}), which can lead to less sample
efficient implementation when compared to the baseline, but the higher FPS
largely compensates for this.  Extending the batch size in the temporal
dimension (N-steps bootstrapping, $N=20$) increases the GPU computational load
and reduces both the FPS and UPS, but it also reduces the bias in the estimate
of the value function, making each DNN update more effective, and leads to an
overall decrease of the wall clock training time, the fastest convergence being
achieved by CuLE with 1,200 environments. Using OpenAI Gym with the same
configuration results in a longer training time, because of the lower FPS
generated by CPU emulation.

\paragraph{Generalization for different systems} Table~\ref{tab:different_GPUs}
reports the FPS for the implementations of vanilla DQN, A2C, and PPO, on System
I and II in Table~\ref{tab:systems}. The speed-up in terms of FPS provided by
CuLE is consistent across different systems, different algorithms, and larger
in percentage for a large number of environments.  Different DRL algorithms
achieve different FPS depending on the complexity and frequency of the training
step on the GPU. 

\begin{table*}[th!]
\centering
\caption{Average FPS and min/max GPU utilization during training for Pong
    with different algorithms and using different emulation engines on
    different systems (see Table~\ref{tab:systems}); CuLE consistently leads to
    higher FPS and GPU utilization.}    
\label{tab:different_GPUs}    
    \begin{scriptsize}
    \centering
    \begin{tabular}{cc||cc|cc}
    \toprule
        Algorithm & Emulation engine & \multicolumn{4}{c}{FPS [GPU utilization \%]} \\
        &         & System I [256 envs] & System I [1024 envs] & System II [256 envs] & System II [1024 envs] \\
        \midrule
        & OpenAI   & 6.4K [15-42\%]  & 8.4K [0-69\%] & 10.8K [26-32\%]  & 21.2K [28-75\%] \\
		DQN &CuLE\textsubscript{CPU} & 7.2K [16-43\%]  & 8.6K [0-72\%] & 6.8K [17-25\%]  & 20.8K [8-21\%] \\
        & CuLE & 14.4K [16-99\%] & 25.6K [17-99\%] & 11.2K [48-62\%]  & 33.2K [57-77\%] \\
        \midrule
        & OpenAI   & 12.8K [2-15\%]  & 15.2K [0-43\%] & 24.4K [5-23\%]  & 30.4K [3-45\%]  \\
		A2C & CuLE\textsubscript{CPU} & 10.4K [2-15\%]  & 14.2K [0-43\%] & 12.8K [1-18\%]  & 25.6K [3-47\%] \\
        & CuLE & 19.6K [97-98\%] & 51K [98-100\%]& 23.2K [97-98\%]  & 48.0K [98-99\%]  \\
        \midrule
        & OpenAI & 12K [3-99\%]    & 10.6K [0-96\%] & 16.0K [4-33\%]  & 19.2K [4-62\%]  \\
        PPO & CuLE\textsubscript{CPU} & 10K [2-99\%]    & 10.2K [0-96\%]& 9.2K [2-28\%]  & 18.4K [3-61\%]   \\
        & CuLE & 14K [95-99\%]   & 36K [95-100\%]& 14.4K [43-98\%]  & 28.0K [45-99\%]\\
\bottomrule    
    \end{tabular}    
    \end{scriptsize}
    
    % \IF{Steven I add
    % here some general observation that should be moved to the text
    % - utilization is lower for DQN as we have to alternate between training and
    % generating data, since we are using a single GPU, thus the advantage of
    % using an off-policy algo is not present in this case. Generally speaking it
    % seems that a multi=GPU system is much more suitable for this kind of
    % algorithm, otherwise the single-GPU implementation is detriemantal also for
    % the FPS, which is higher for A2C. The peak performance never reaches 100\%
    % for CPU data engines, as we are limited by bandwidth - while we reach 100
    % for CuLE GPU, as the GPU is always busy generating or processing data. The
    % interval reported here is the min / max utilization I observed. finally,
    % PPO is an algorithm that requires more gradients and so on, it requires too
    % much memory for large batch size.}
    
\end{table*}

\section{Conclusion}
~\label{sec:conclusion}

%[Mention use of correlation and state seeding to increase FPS?]

%[HERE??? Overall, our analysis shows once more that, compared to OpenAI Gym, CuLE
%guarantees faster and more stable convergence; the batching strategy plays
%a significant role in this, as CuLE allows splitting a large set of
%environments into large batches, that guarantee a low variance in the estimate
%of the value, that are also long in the temporal direction, that guarantee
%small bias.  This eventually leads to faster experimental turnaround time
%during the development and analysis of existing and novel DRL algorithms.]

% The common allocation of the tasks in a DRL system dictates that the environment
% should run on CPUs, whereas GPUs should be dedicated to DNN operations.  With
% most of the existing frameworks~\cite{OpenAIGym, Mirowski:2016, Tassa:2018,
% Tian:2017} following this paradigm, the limited CPU-GPU communication bandwidth
% and CPU capability to emulate a large number of environments represent two
% limiting factors to effectively accelerate DRL algorithms, even when mapped to
% expensive distributed systems.  By rendering frames directly on the GPU, CuLE
% overcomes these limitations and  generate as many FPS as those generated by
% large, expensive CPU systems.  CuLE promises to be an effective tool to develop
% and test DRL algorithms by significantly reducing the experiment turnaround
% time.

As already shown by others in the case of DRL on distributed system, our
experiments show that proper batching coupled with a slight off-policy gradient
policy algorithm can significantly accelerate the wall clock convergence time;
CuLE has the additional advantage of allowing effective scaling of this
implementation to a system with multiple GPUs.
%Our analysis further highlights that CuLE generates training frames with
%a pattern that is distinct from data generated by CPU emulators: since GPUs
%are characterized by a number of instructions per second perthread which is
%lower compared to that of a CPU, CuLE does achieve an higher FPS, but FPS
%/ environment is higher forthe CPU.  
CuLE effectively allows increasing the number of parallel environments but,
because of the low number of instructions per second per thread on the GPU,
training data can be narrow in the time direction.  This can be problematic for
problems with sparse temporal rewards,
% solved by a naive DRL implementation,
but rather than considering this as a pure limitation of CuLE, we believe that
this peculiarity opens the door to new interesting research questions, like
active sampling of important states~\cite{Hessel:2017,Wang:2016} that can then
be effectively analyzed on a large number of parallel environments with CuLE.
CuLE also hits a new obstacle, which is the limited amount of DRAM available on
the GPU; studying new compression schemes, like the one proposed
in~\cite{Hessel:2017}, as well as training methods with smaller memory
footprints may help extend the utility of CuLE to even larger environment
counts, and design better GPU-based simulator for RL in the future.  Since
these are only two of the possible research directions for which CuLE is an
effective investigation instrument, CuLE comes with a python interface that
allows easy experimentation and is freely available to any researcher at
\culeurl.

% A last note has to be done on CuLE's implementation, that is informed by ease
% of debugging, need for flexibility, and compatibility with standard DRL
% benchmarks.  These choices put a limit on the achievable speed up factor (for
% instance by using emulation of the Atari 2600 console instead of direct CUDA
% implementations of Atari games), but the analysis and insights provided in our
% paper furnish indications for the design of efficient simulators for DRL.

\pagebreak
\section{Impact Statement}~\label{sec:impact_statement}

As interest in deep reinforcement learning has grown so has the computational
requirements for researchers in this field.  However, the reliance of DRL on
the CPU, especially for environment simulation/emulation, severely limits the
utilization of the computational resources typically accessible to DL
researchers, specifically GPUs.  Though Atari is a specialized DRL environment,
it is arguably one of the most studied in recent times and provides access to
several training environments with various levels of difficulty. The
development and testing of DRL using Atari games remains a relevant and
significant step toward more efficient algorithms. There are two impact points
for CuLE: 1) Provide access to an accelerated training environment to
researchers with limited computational capabilities. 2) Facilitate research in
novel directions that explore thousands of agents without requiring access to
a distributed system with hundreds of CPU cores.  Although leaving RL
environments "as-is" on CPUs and parallelizing across multiple nodes is indeed
the shortest path to make progress is it also inherently inefficient, in terms
of the resource utilization on the local machine, and expensive, since it
requires access to a large number of distributed machines. The more efficient
use of the computational resources could also lead to a smaller carbon
footprint.

% Traditional DRL training focuses on CPU environments that execute a set of
% actions $\{a_{t-1}\}$ at time $t-1$, and produce observable states $\{s_t\}$
% and rewards $\{r_t\}$.  These data are migrated to a Deep Neural Network (DNN)
% on the GPU to eventually select the next action set, $\{a_t\}$, which is copied
% back to the CPU.  This sequence of operations defines the \emph{inference
% path}, whose main aim is to generate training data.  A training buffer on the
% GPU stores the states generated on the \emph{inference path}; this is
% periodically used to update the DNN's weights $\theta$, according to the
% training rule of the DRL algorithm (\emph{training path}).  A computationally
% efficient DRL system should balance the data generation and training processes,
% while minimizing the communication overhead along the \emph{inference path} and
% consuming, along the \emph{training path}, as many data per second as
% possible~\cite{Babaeizadeh:2016,Babaeizadeh:2017}.  The solution to this
% problem is however non-trivial and many DRL implementations do not leverage the
% full computational potential of modern systems~\cite{Stooke:2018}.

\newcommand{\beq}{\begin{equation}}
\newcommand{\eeq}{\end{equation}}
\newcommand{\beqa}{\begin{eqnarray}}
\newcommand{\eeqa}{\end{eqnarray}}
\newcommand{\beqan}{\begin{eqnarray*}}
\newcommand{\eeqan}{\end{eqnarray*}}
\renewcommand{\P}{\mathbb{P}}
\renewcommand{\Pr}{\mathbb{P}}
\newcommand{\Q}{\mathbb{Q}}
\newcommand{\Esp}{\mathbb{E}}
\newcommand{\indic}[1]{\mathbb{I}\{#1\}}
\newcommand{\EE}[1]{\E\left[#1\right]}
\newcommand{\Prob}[1]{\P\left(#1\right)}
\newcommand{\wh}{\widehat}
\newcommand{\eqdef}{\stackrel{\rm def}{=}}

\section{Appendix}

\subsection{Reinforcement Learning, A2C and V-trace}

\paragraph{Reinforcement learning} In RL, an agent observes a state $s_t$ at
time $t$ and follows a policy $\pi=\pi(s_t)$ to select an action $a_t$; the
agent also receives a scalar reward $r_t$ from the environment.  The goal of RL
is to optimize $\pi$ such that the sum of the expected rewards is maximized.

In model-free policy gradient methods $\pi(a_t|s_t;\theta)$ is the output of
a policy DNN with weights $\theta$, and represents the probability of selecting
action $a_t$ in the state $s_t$.  Updates to the DNN are generally aligned in
the direction of the gradient of $E[R_t]$, where
$R_t=\sum_{i=0}^{\infty}\gamma^ir_{t+i}$ is the discounted reward from time
$t$, with discount factor $\gamma\in (0, 1]$ (see also
REINFORCE~\cite{williams1992simple}) The vanilla implementation updates
$\theta$ along $\nabla_\theta\log\pi(a_t|s_t;\theta)R_t$, which is an unbiased
estimator of $\nabla_\theta E[R_t]$.  The training procedure can be improved by
reducing the variance of the estimator by subtracting a learned
\textit{baseline} $b_t(s_t)$ and using the gradient
$\nabla_\theta\log\pi(a_t|s_t;\theta)[R_t~-~b_t(s_t)]$.  One common baseline is
the value function $V^\pi(s_t) = E[{R_t|s_t}]$, which is the expected return
for the policy $\pi$ starting from $s_t$.  The policy $\pi$ and the baseline
$b_t$ can be viewed as \textit{actor} and \textit{critic} in an actor-critic
architecture \cite{Sutton:1998:IRL:551283}.

\paragraph{A2C} A2C~\cite{openaiblog} is the synchronous version of
A3C~\cite{Mnih:2016}, a successful actor-critic algorithm, where a single DNN
outputs a softmax layer for the policy $\pi\left(a_t | s_t; \theta \right)$,
and a linear layer for $V \left(s_t; \theta\right)$.  In A2C, multiple agents
perform simultaneous steps on a set of parallel environments, while the DNN is
updated every $t_{max}$ actions using the experiences collected by all the
agents in the last $t_{max}$ steps.  This means that the variance of the critic
$V\left(s_t; \theta\right)$ is reduced (at the price of an increase in the
bias) by $N$-step bootstrapping, with $N = t_{max}$.  The cost function for the
policy is then:
\begin{equation}
%f_{\pi}\left( \theta \right)=
 \log\pi\left(a_t | s_t; \theta \right) \left[\tilde{R}_t - V\left(s_t;
\theta_t \right) \right] + \beta H \left[\pi \left(s_t; \theta \right)\right],
\label{eq:costPi} \end{equation}
where $\theta_t$ are the DNN weights $\theta$ at time $t$, $\tilde{R}_t
= \sum_{i=0}^{k-1}{\gamma^i r_{t+i} + \gamma^k V\left(s_{t+k}; \theta_t
\right)}$ is the bootstrapped discounted reward from $t$ to $t+k$ and $k$ is
upper-bounded by $t_{max}$, and $H\left[\pi \left(s_t; \theta \right)\right]$
is an entropy term that favors exploration, weighted by the hyper-parameter
$\beta$.  The cost function for the estimated value function is:
\begin{equation}
%f_v \left(\theta\right)=
\left[\tilde{R}_t - V \left( s_t; \theta \right) \right]^2, \label{eq:costV}
\end{equation}
which uses, again, the bootstrapped estimate $\tilde{R}_t$. Gradients $\nabla
\theta$ are collected from both of the cost functions; standard optimizers, such as
Adam or RMSProp, can be used for optimization.

\paragraph{V-trace} In the case where there is a large number of environments,
such as in CuLE or IMPALA~\cite{Espeholt:2018}, the synchronous nature of A2C
become detrimental for the learning speed, as one should wait for all the
environments to complete $t_{max}$ steps before computing a single DNN update.
Faster convergence is achieved (both in our paper and in~\cite{Espeholt:2018})
by desynchronizing data generation and DNN updates, which in practice means
sampling a subset of experiences generated by the agents, and updating the
policy using an approximate gradient, which makes the algorithm slightly
off-policy.

To correct for the off-policy nature of the data, that may lead to inefficiency
or, even worse, instabilities, in the training process, V-trace is introduced
in~\cite{Espeholt:2018}.  In summary, the aim of off-policy correction is to
give less weight to experiences that have been generated with policy $\mu$,
called the {\em behaviour policy}, when it differs from the {\em target
policy}, $\pi$; for a more principled explanation we remand the curios reader to
~\cite{Espeholt:2018}.

For a set of experiences collected from time $t=t_{0}$ to time $t=t_{0}+N$ following
some policy $\mu$, the $N$-steps V-trace target for $V(s_{t_{0}}; \theta)$ is
defined as:

\begin{eqnarray}v_{t_{0}} & = & \textstyle{V(s_{t_{0}}; \theta) + \sum_{t=t_{0}}^{t_{0}+N-1} \gamma^{t-t_{0}} \Big( \prod_{i=t_{0}}^{t-1} c_i \Big ) \delta_t V}, \label{eq:target.off}\\
\delta_t V & = & \rho_t \big(r_t+\gamma V(s_{t+1}; \theta)-V(s_t; \theta)\big) \\
\rho_t & = & \min\big(\bar\rho, \frac{\pi(a_t|s_t)}{\mu(a_t|s_t)}\big)\label{eq:rho}\\
c_i & = & \min\big(\bar c, \frac{\pi(a_i|s_i)}{\mu(a_i|s_i)}\big)\label{eq:c};
\end{eqnarray}

$\rho_t$ and $c_i$ are truncated importance sampling (IS) weights, and
$\prod_{i=t0}^{t-1} c_i=1$ for $s=t$, and $\bar \rho\geq \bar c$.  Notice that,
when we adopt the proposed multi-batching strategy, there are multiple behaviour
policies $\mu$ that  have been followed to generate the training data --- e.g.,
N different policies are used when SPU=1 in Fig.~\ref{fig:vtrace}.
Eqs.~\ref{eq:rho}-\ref{eq:c} do not need to be changed in this case, but we
have to store all the $\mu(a_i|s_i)$ in the training buffer
%in Fig.~\ref{fig:design}
to compute the, V-trace corrected, DNN update.
In our implementation, we compute the V-trace update recursively as:
\begin{equation}
v_t = V(s_t; \theta) + \delta_tV + \gamma c_s\big( v_{t+1} - V(s_{t+1}; \theta)\big).
\end{equation}

%Now in the off-policy setting that we consider, we can use an IS weight
%between the policy being evaluated $\pi_{\bar\rho}$ and the behaviour policy
%$\mu$, to update our policy parameter in the direction of
%\beq\label{eq:off-policy.PG}
%\E_{a_s\sim \mu(\cdot|x_s)}\Big[ \frac{\pi_{\bar \rho}(a_s|x_s)}{\mu(a_s|x_s)} \nabla\log\pi_{\bar \rho}(a_s|x_s) q_s \big| x_s\Big]
%\eeq
%where $q_s \eqdef r_s + \gamma v_{s+1}$ is an estimate of  $Q^{\pi_{\bar \rho}}(x_s, a_s)$ built from the V-trace estimate $v_{s+1}$ at the next state $x_{s+1}$.
%
%In order to reduce the variance of the policy gradient
%estimate~\eqref{eq:off-policy.PG}, we usually subtract from $q_s$
%a state-dependent baseline, such as the current value approximation $V(x_s)$.
%
%Finally notice that \eqref{eq:off-policy.PG} estimates the policy gradient for
%$\pi_{\bar \rho}$ which is the policy evaluated by the V-trace algorithm when
%using a truncation level $\bar \rho$. However assuming the bias
%$V^{\pi_{\bar\rho}}-V^{\pi}$ is small (e.g.~if $\bar\rho$ is large enough)
%then we can expect $q_s$ to provide us with a good estimate of
%$Q^{\pi}(x_s,a_s)$. Taking into account these remarks, we derive the following
%canonical V-trace actor-critic algorithm.

At training time $t$, we update $\theta$ with respect to the value output, $v_s$, given by:
\begin{equation}
\big(v_t - V(s_t; \theta)\big) \nabla_\theta V(s_t; \theta),
\end{equation}
whereas the policy gradient is given by:
\begin{equation}
\rho_t \nabla_\omega\log\pi_\omega(a_s|s_t) \big( r_t+\gamma v_{t+1} - V(s_t; \theta)\big).
\end{equation}
An entropy regularization term that favors exploration and prevents premature convergence (as in Eq.~\ref{eq:costPi}) is also added.

~\begin{figure*}[]
\centering
  \subfigure[Asterix, CPU]{\includegraphics[width=0.31\textwidth]{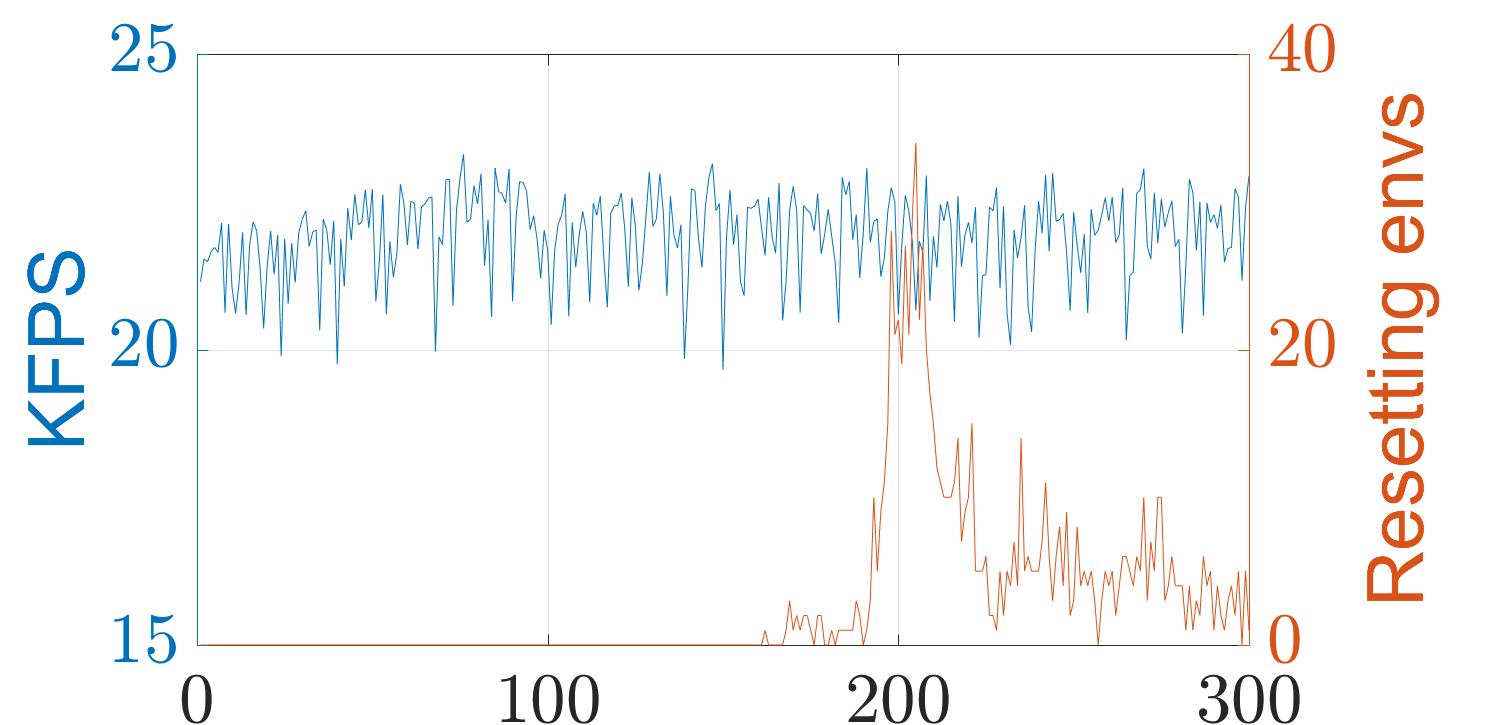}~\label{fig:decor_asterix_cpu}}
  \subfigure[Pong, CPU]{\includegraphics[width=0.31\textwidth]{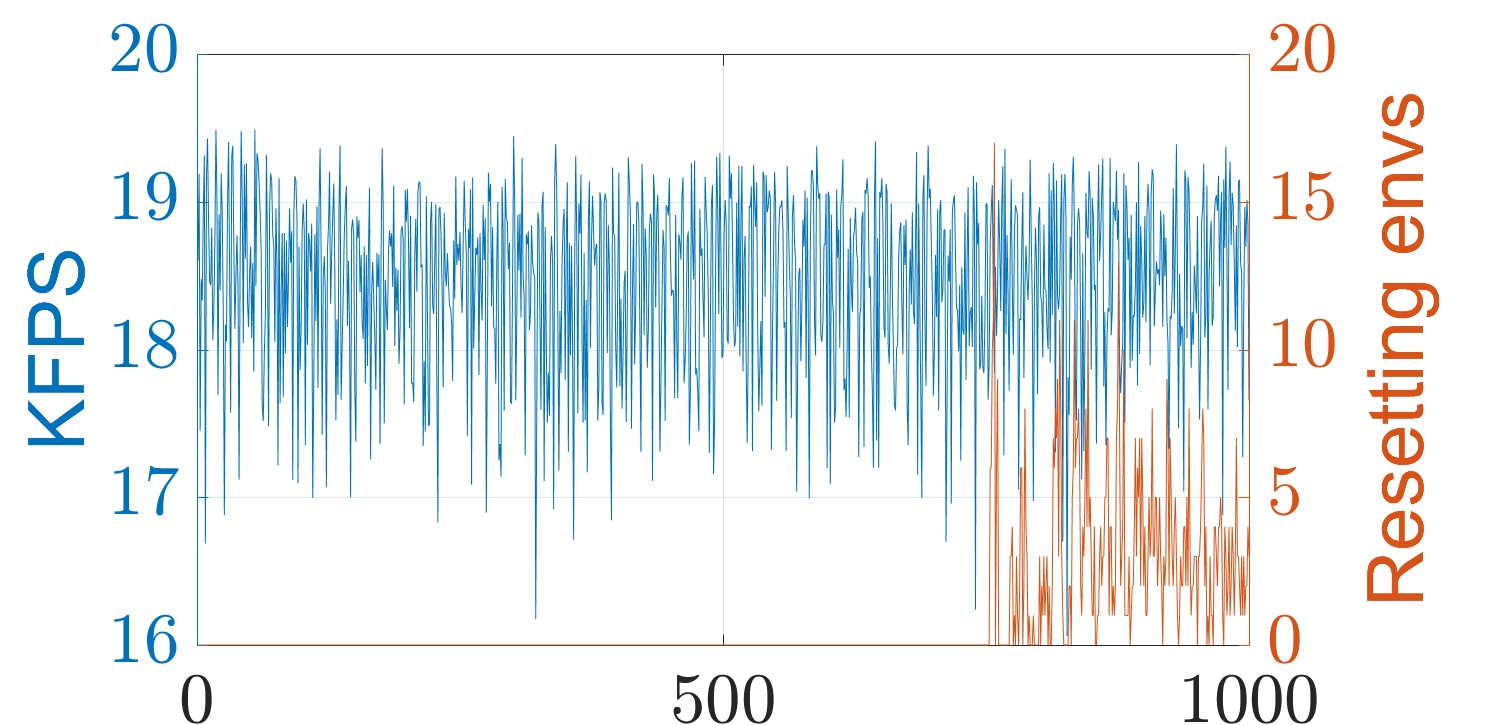}~\label{fig:decor_pong_cpu}}\\
  \subfigure[Ms Pacman, CPU]{\includegraphics[width=0.31\textwidth]{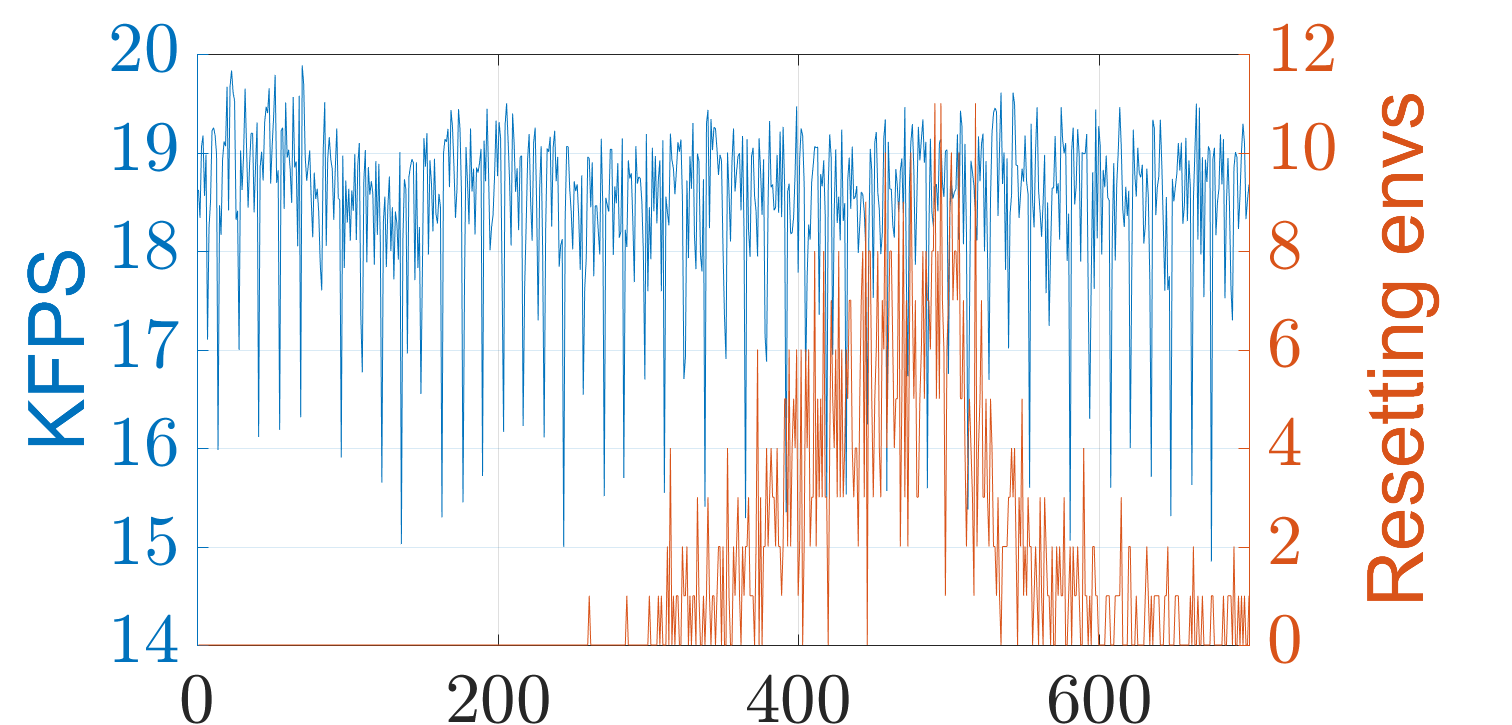}~\label{fig:decor_mspacman_cpu}}
  \subfigure[Assault, CPU]{\includegraphics[width=0.31\textwidth]{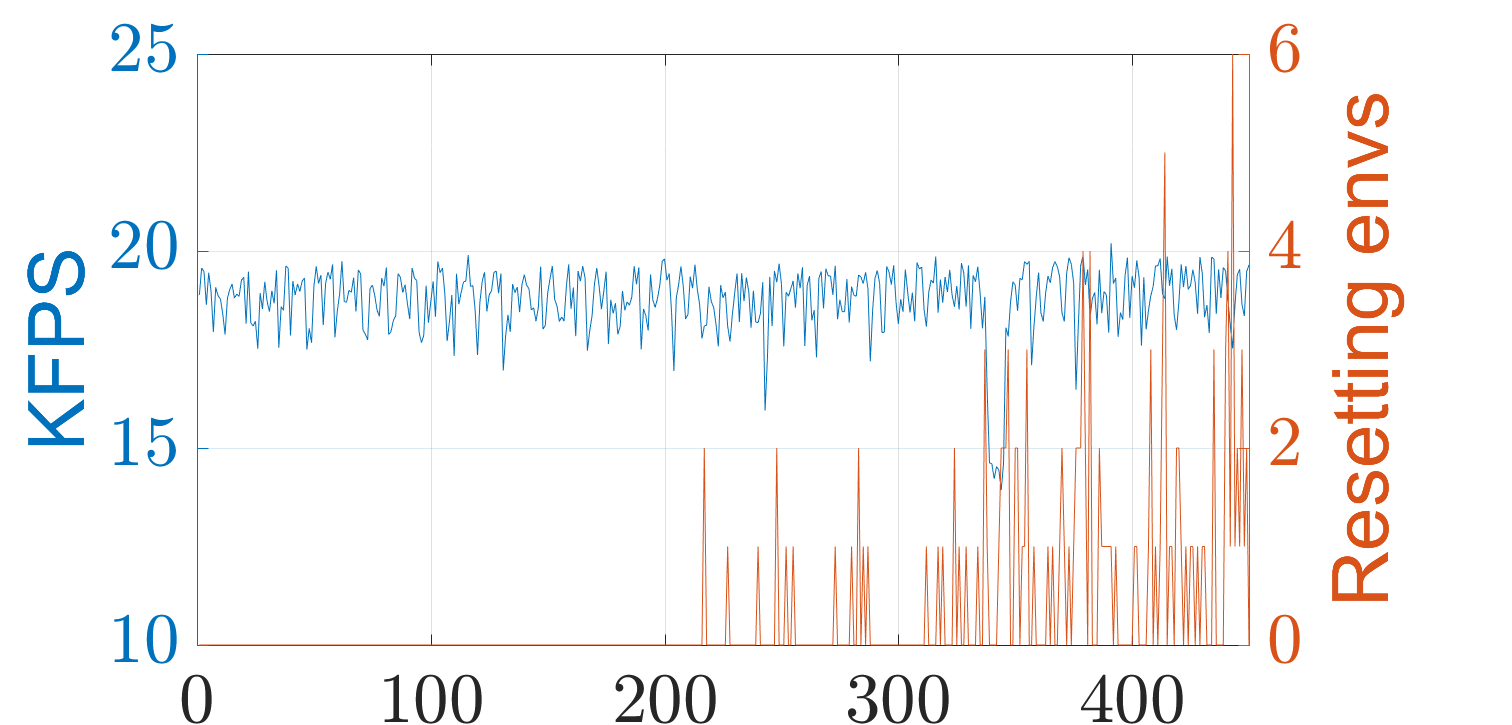}~\label{fig:decor_assault_cpu}}
\caption{FPS as a function of the environment step, measured on System I in
  Table~\ref{tab:systems} for \emph{emulation only} on four Atari games, 512
  environments, for CuLE\textsubscript{CPU}; each panel also shows the number
  of resetting environments.  A peak in the FPS at the beginning of the
  emulation period, as in the case of GPU emulation in Fig.~\ref{fig:decor}, is
  not visible in this case.}
\label{fig:decor_cpu}
\end{figure*}

\subsection{Thread divergence is not present in the case of CPU emulation}

We show here that thread divergence, that affects GPU-based emulation (see
Fig.~\ref{fig:decor}), does not affect CPU-based emulation.
Fig.~\ref{fig:decor_cpu} shows the FPS on four Atari games where all the
environments share the same initial state.  In constrast with GPU
emulation, the CPU FPS do not peak at the beginning of the emulation period, where
many environments are correlated.

\subsection{Performance during training - other games}

\begin{figure*}[]
\centering
  \subfigure[Pong]{\includegraphics[width=0.32\textwidth]{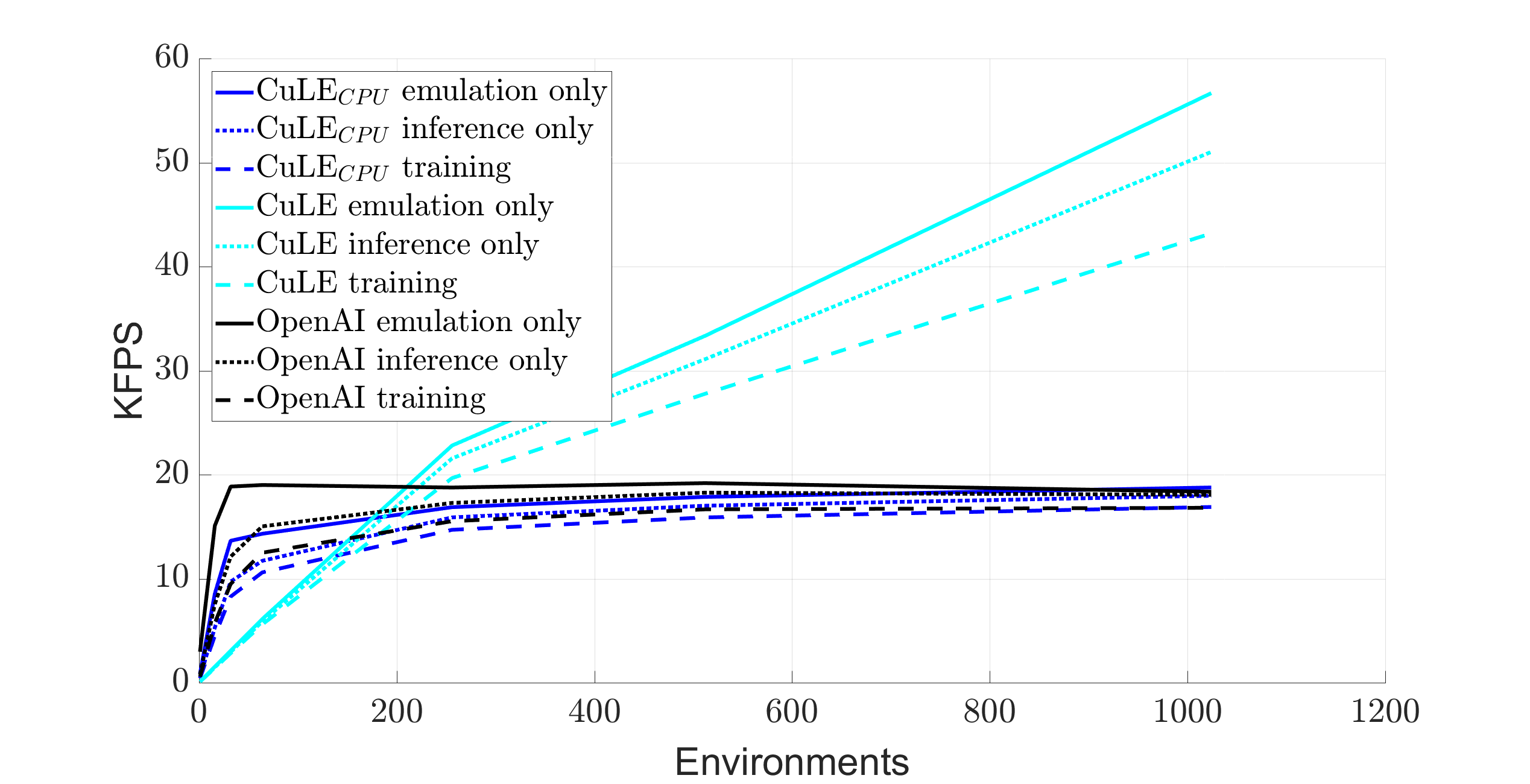}~\label{fig:bench_pong}}
  \subfigure[MsPacman]{\includegraphics[width=0.32\textwidth]{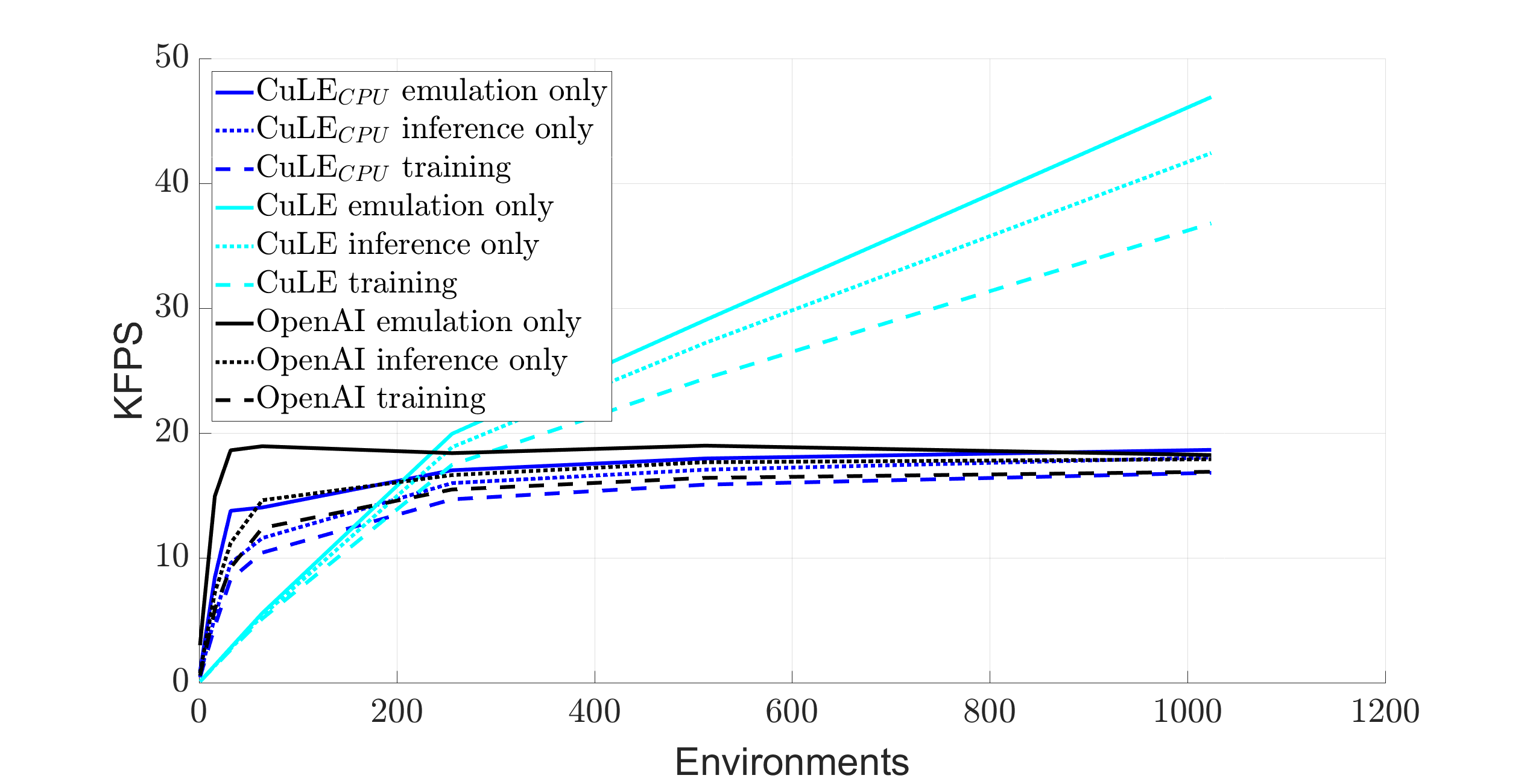}~\label{fig:bench_mspacman}}
  \subfigure[Asterix]{\includegraphics[width=0.32\textwidth]{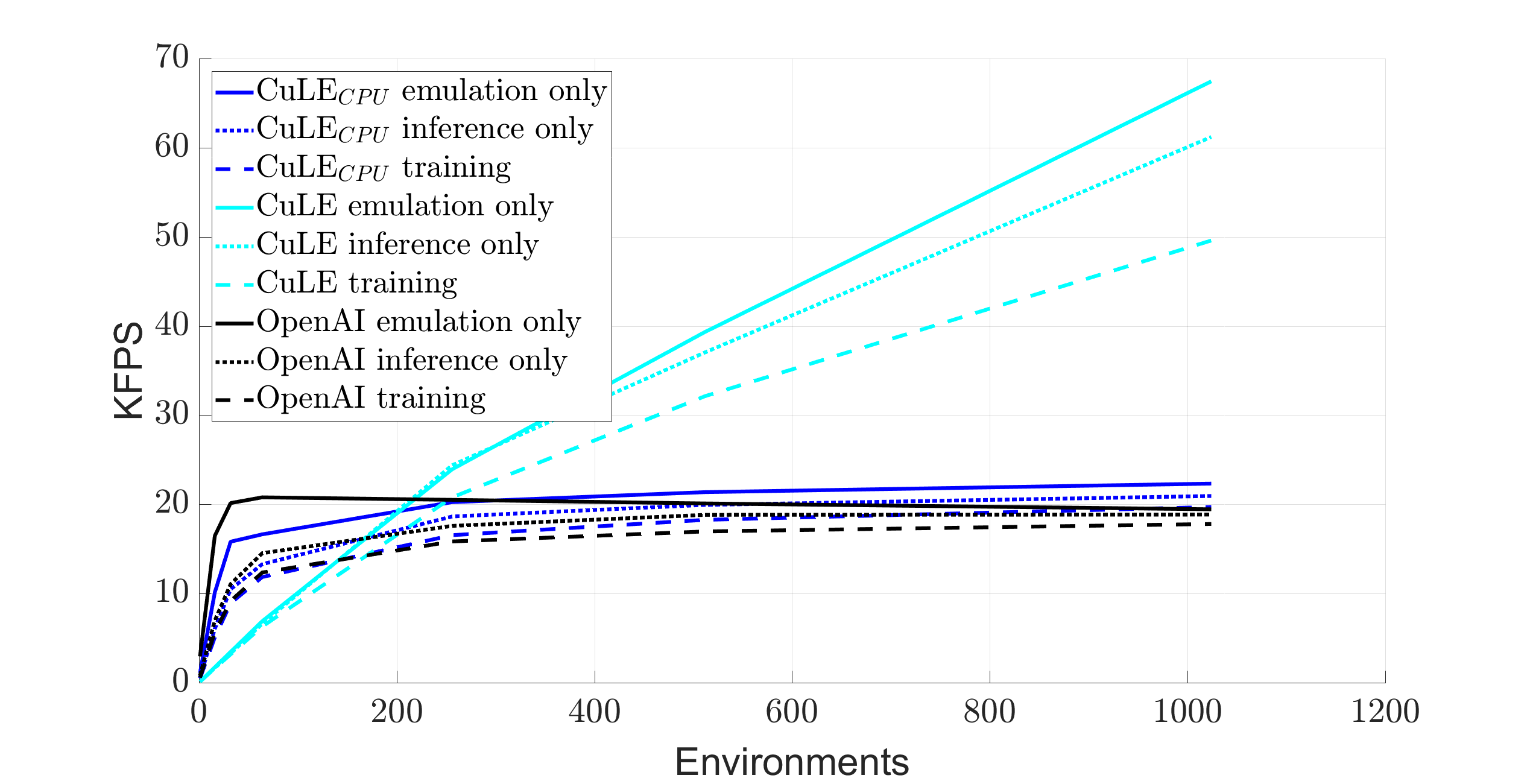}~\label{fig:benasterix}}
\caption{FPS generated by different emulation engines on System I in
  Table~\ref{tab:systems} for different Atari games, as a function of the
  number of environments, and different load conditions (the main
  A2C~\cite{openaiblog} loop is run here, with N-step bootstrapping, $N=5$.}
  \label{fig:bench_training_appendix}
\end{figure*}

For sake of space, we only report (Fig.~\ref{fig:bench_training_appendix}) the
FPS measured on system I in Table~\ref{tab:systems} for three additional games,
as a function of different load conditions and number of environments.

~

\subsection{Correctness of the implementation}

\begin{figure*}[]
\centering
  \subfigure[Assault]{\includegraphics[width=0.23\textwidth]{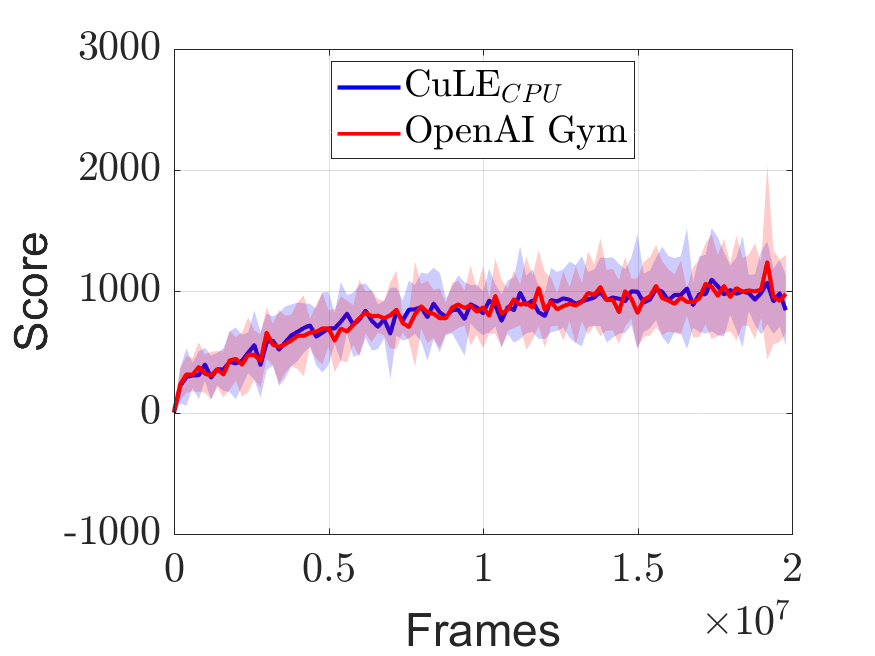}~\label{fig:corr_assault}}
  \subfigure[Asterix]{\includegraphics[width=0.23\textwidth]{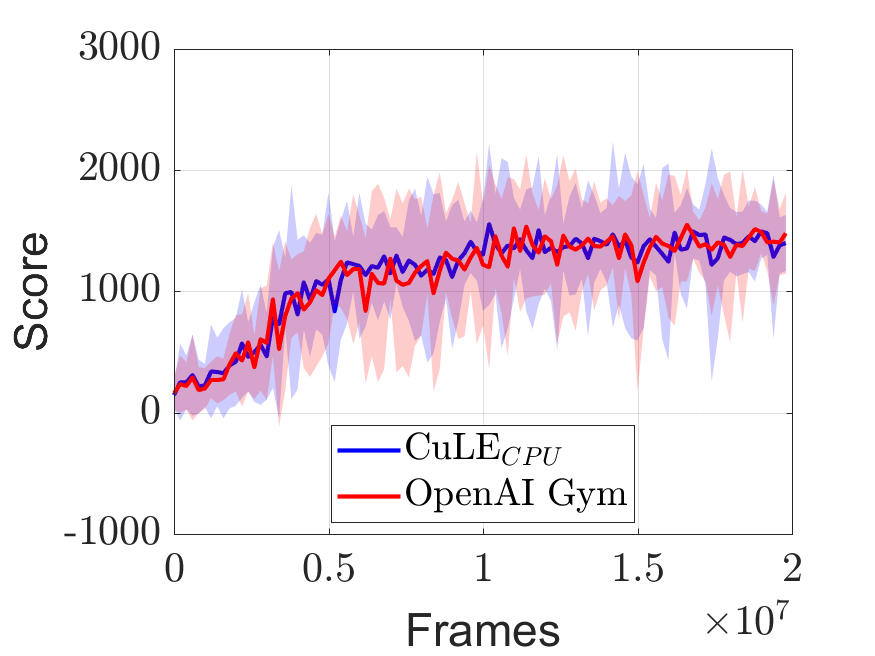}~\label{fig:corr_asterix}}
  \subfigure[Ms-Pacman]{\includegraphics[width=0.23\textwidth]{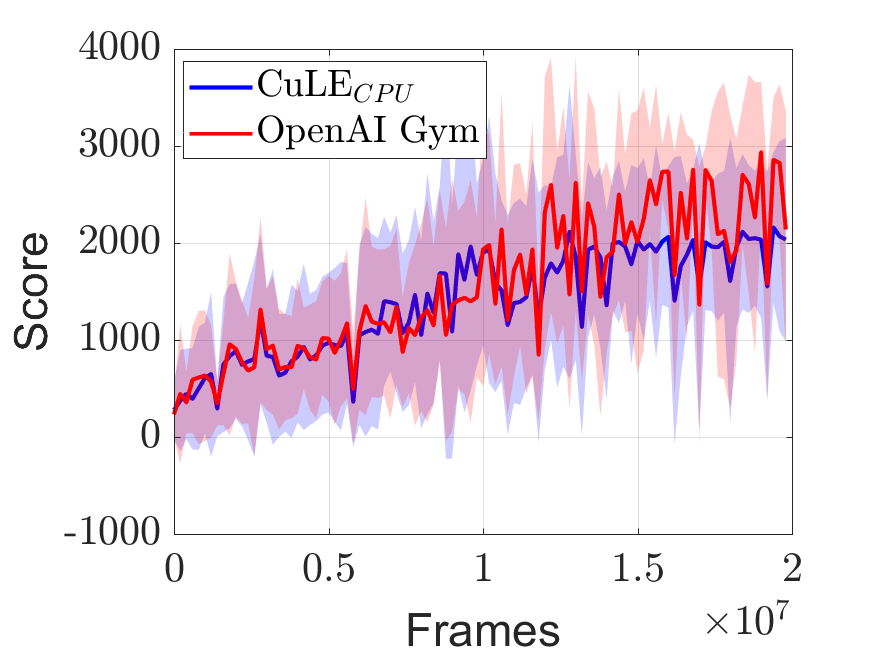}~\label{fig:corr_pacman}}
  \subfigure[Pong]{\includegraphics[width=0.23\textwidth]{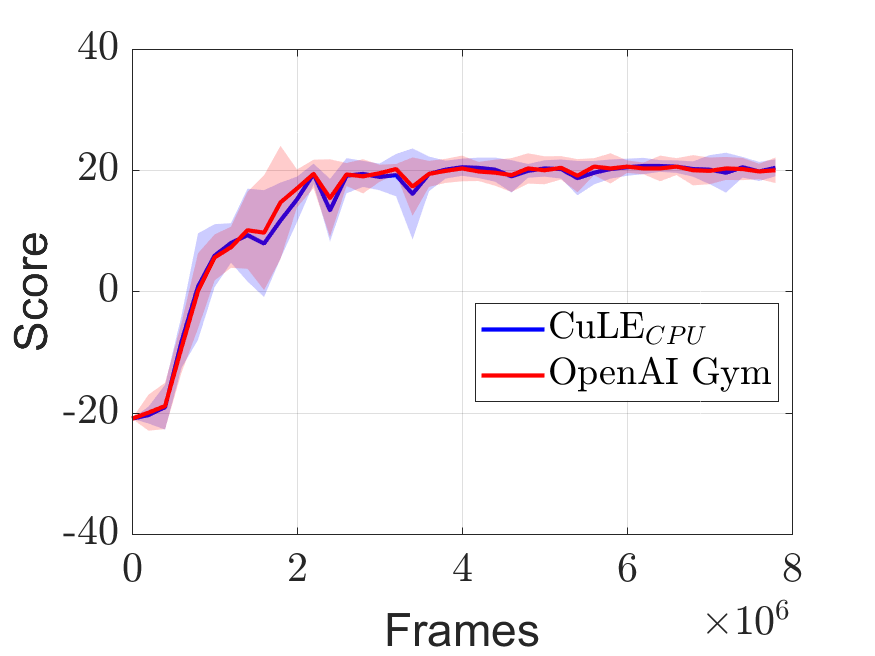}~\label{fig:corr_pong}}
  \caption{Average testing scores measured on 10 CuLE\textsubscript{CPU} and
  OpenAI Gym environments, while training with A2C+V-trace and CuLE, as
  a function of the training frames; 250 environments are used for Ms-Pacman,
  given its higher variability. The shaded area represents 2 standard
  deviations.}
  \label{fig:correctness}
\end{figure*}

To demonstrate the correctness of our implementation, and thus that policies
learned with CuLE generalize to the same game emulated by OpenAI Gym, we report
in Fig.~\ref{fig:correctness} the average scores achieved in testing, while
training an agent with with A2C+V-trace and CuLE. The testing scores measured
on CuLE\textsubscript{CPU} and OpenAI Gym environments do not show any relevant
statistical difference, even for the case of Ms-Pacman, where the variability
of the scores is higher because of the nature of the game.

\bibliography{Arxiv_GPURL}
\bibliographystyle{plain}

\end{document}